%% file: main.tex
\documentclass[sigconf]{acmart}
\AtBeginDocument{%
  \providecommand\BibTeX{{%
    \normalfont B\kern-0.5em{\scshape i\kern-0.25em b}\kern-0.8em\TeX}}}

\usepackage{ulem}
\usepackage{graphicx}
\usepackage{array}
\usepackage{booktabs}
\usepackage{multicol,lipsum}
\usepackage{cancel}
\usepackage{caption}
\usepackage{wrapfig}
\usepackage{tabularx}
\usepackage{multirow}
\usepackage{gensymb}

\newcolumntype{C}[1]{>{\centering\arraybackslash}m{#1}}
\renewcommand{\arraystretch}{1} 

\copyrightyear{2024}
\acmYear{2024}
\setcopyright{acmlicensed}\acmConference[HRI '24]{Proceedings of the 2024 ACM/IEEE International Conference on Human-Robot Interaction}{March 11--14, 2024}{Boulder, CO, USA}
\acmBooktitle{Proceedings of the 2024 ACM/IEEE International Conference on Human-Robot Interaction (HRI '24), March 11--14, 2024, Boulder, CO, USA}
\acmDOI{10.1145/3610977.3634989}
\acmISBN{979-8-4007-0322-5/24/03}

\begin{document}

\title{RABBIT: A Robot-Assisted Bed Bathing System with Multimodal Perception and Integrated Compliance}

\author{Rishabh Madan}
\authornote{Both authors contributed equally to this research.}
\author{Skyler Valdez}
\authornotemark[1]
\email{rm773@cornell.edu}
\email{skyleravaldez@gmail.com}
\affiliation{%
  \institution{Cornell University}
  \city{Ithaca}
  \state{NY}
  \country{USA}
}

\author{David Kim}
\author{Sujie Fang}
\author{Luoyan Zhong}
\author{Diego T. Virtue}
\affiliation{%
  \institution{Cornell University}
  \city{Ithaca}
  \state{NY}
  \country{USA}
}

\author{Tapomayukh Bhattacharjee}
\email{tapomayukh@cornell.edu}
\affiliation{%
  \institution{Cornell University}
  \city{Ithaca}
  \state{NY}
  \country{USA}
}






\renewcommand{\shortauthors}{Rishabh Madan, et al.}

\input{src/abstract}



\maketitle

\input{src/introduction}
\input{src/related-work}
\input{src/methods}
\input{src/experiments}
\input{src/user-study}
\input{src/discussion}

\section{Acknowledgements}
This work was partly funded by NSF IIS \#2132846, CAREER \#2238792, and DARPA under Contract HR001120C0107.

\bibliographystyle{ACM-Reference-Format}
\bibliography{sample-base}

\appendix

\end{document}

%% file: src/abstract.tex
\begin{abstract}
This paper introduces RABBIT, a novel robot-assisted bed bathing system designed to address the growing need for assistive technologies in personal hygiene tasks. It combines multimodal perception and dual (software and hardware) compliance to perform safe and comfortable physical human-robot interaction. Using RGB and thermal imaging to segment dry, soapy, and wet skin regions accurately, RABBIT can effectively execute washing, rinsing, and drying tasks in line with expert caregiving practices. Our system includes custom-designed motion primitives inspired by human caregiving techniques, and a novel compliant end-effector called Scrubby, optimized for gentle and effective interactions. We conducted a user study with 12 participants, including one participant with severe mobility limitations, demonstrating the system's effectiveness and perceived comfort. Supplementary material and videos can be found on our website \href{https://emprise.cs.cornell.edu/rabbit}{\textbf{emprise.cs.cornell.edu/rabbit}}.
\end{abstract}

%% file: src/introduction.tex
\section{Introduction}

A 2021 survey revealed that over a billion individuals worldwide needed assistance with activities of daily living (ADLs), underscoring the growing importance of caregiving \cite{abdi2021emerging}. One such basic ADL is bathing, a critical component of maintaining personal hygiene \cite{dunlop1997disability}. For those with severe mobility limitations, traditional bathing methods may not be feasible, necessitating bed bathing as an alternative \cite{downey2008bed}. Caregiving robots have the potential to provide assistance with such basic ADLs and improve the quality of life for care recipients while reducing caregiver burden \cite{darragh2015musculoskeletal, czuba2012ergonomic}.

\input{figures/src/wow_figure}
Robots offer a myriad of advantages, from precision and repeatability to the ability to operate tirelessly. However, the application of robots to intricate physical human-robot interaction (pHRI) tasks, such as bed bathing, is not straightforward. While there has been commendable progress in robotics, including manipulating deformable objects like towels \cite{zhu2022challenges, matas2018sim, sanchez2018robotic}, interpreting body poses of individuals on beds \cite{liu2020simultaneously, liu2022simultaneously}, and ensuring unobtrusive movement around a human \cite{jakopin2017unobtrusive}, several gaps remain. Most notably, existing robot-assisted bed bathing approaches \cite{zlatintsi2017, Dometios2017, Yukawa2010, Huang2022} fall short when it comes to implementing caregiving practices that are endorsed and recommended by clinical experts, such as clinical nurse specialists and occupational therapists (OTs). For instance, research conducted by \cite{Huang2022, king2015} purely focused on clearing debris but did not account for commonly performed tasks during bed bathing, such as applying soap or drying wet regions of the body. 

Human caregivers, with their expertise in bed bathing, perform this task systematically \cite{CareChannel_2019, albertacare} by identifying the differences between the dry, soapy, and wet regions of the skin and modulating their techniques accordingly. For instance, when applying soap, they usually employ linear wiping motions along limbs and then transition to pat drying techniques, especially for people with skin sensitivities or other underlying medical conditions.

Translating these intricate caregiving practices into a robotic system presents multifaceted technical challenges. The first challenge is the image segmentation task, which is complicated by the diverse tones and textures of human skin as well as the translucency of water and its variable temperatures. While there has been progress in detecting transparent objects, the difficulty of identifying suitable sensor modalities for this specific problem, coupled with the absence of a domain-specific dataset, further exacerbates this challenge. The second challenge pertains to motion planning. The robot must not only determine the correct task but also generate trajectories that replicate the motions employed by human caregivers. Lastly, the physical interaction between the robot and the individual is of utmost importance. This requires a system that incorporates both software and hardware compliance to adapt to the uneven contours and variable rigidity of the human body. The tool used for bathing must also be compliant, increasing adaptability to limb contours, especially in the presence of noisy force sensor readings, to further enhance user comfort.

We propose RABBIT, a robot-assisted bed bathing system that uses multimodal perception blended with software and hardware compliance to provide a safe and comfortable bed bathing experience. To be able to train state-of-the-art segmentation models that work well for our problem, we design a novel data collection scheme that allows us to collect realistic data using a nursing manikin. We implement force and position-parameterized motion primitives to generate motion plans for treating the segmented area based on professional caregiving practices. Finally, we design a custom end-effector made out of springs and mounted with soft sponges, to provide hardware compliance and implement a gain-scheduling task-space compliance controller \cite{Khatib1987, Kim2019} to ensure compliance at the software level. We summarize our contributions as follows:

\begin{itemize}
\vspace{-3pt}
    \item We develop a robot-assisted bed bathing system that uses RGB and thermal-based multimodal perception, motion primitives inspired by caregiving practices, and dual (software and hardware) compliance to execute these actions in an efficient and comfortable manner.
    \item We release the Synthetic Bathing Perception (SBP) dataset \cite{rabbit23}, a collection of over 1000 RGB and Thermal images with variations in camera pose, water temperature, presence of soap, presence of hair on the arm, and skin tone.
    \item We benchmark the performance of various image segmentation models that perform RGB and RGB-Thermal image segmentation for our dataset.
    \item We demonstrate the effectiveness and perceived comfort of our system in performing the various tasks within bed bathing through a user study comprising 11 participants with no mobility limitations and one participant with severe mobility limitations.
    \item We release "Scrubby", an open-source custom-designed end-effector comprising springs and soft sponges optimized for gentle and effective bathing interactions.
\end{itemize}
With systems like RABBIT, we aim to bridge the gap between the capabilities of advanced robotics and the nuanced requirements of caregiving. By developing a system that draws inspiration from human caregivers and addresses the unique challenges of bed bathing, we not only aim to enhance the quality of life for care recipients but also pave the way for future research in this domain.

%% file: figures/src/wow_figure.tex
\begin{figure}[t!]
      {\includegraphics[width=1.0\columnwidth]{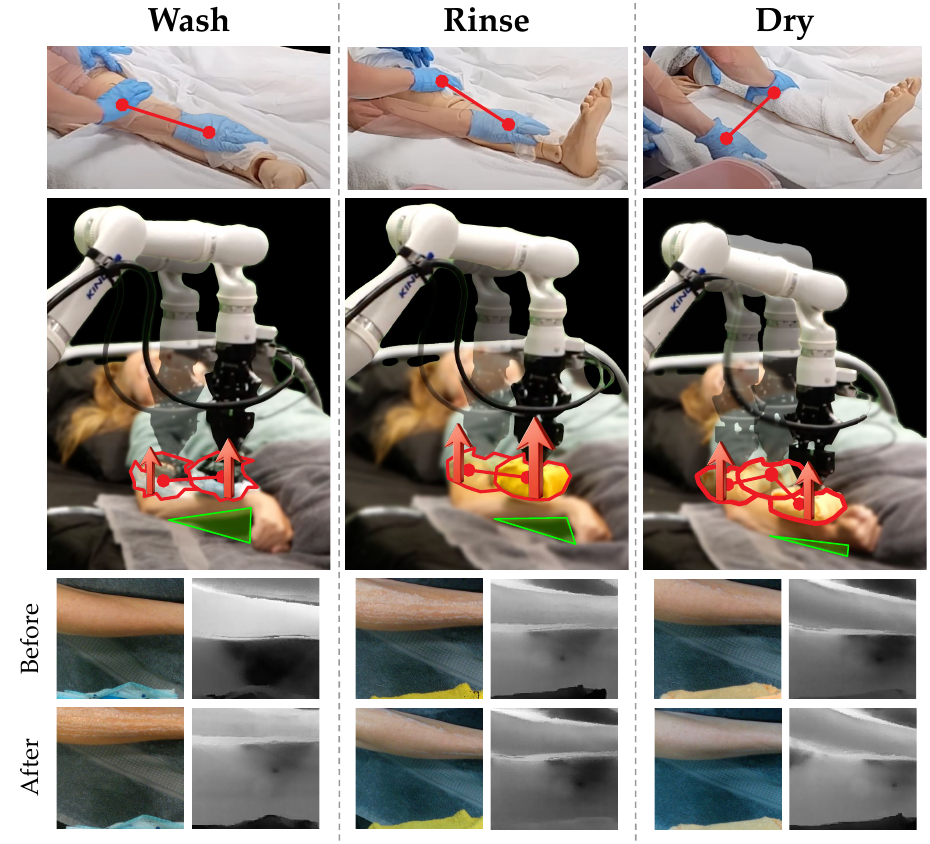}}
  \vspace{-12pt}
  \caption{RABBIT is a robot-assisted bed bathing system informed by expert caregiving practices. By effectively executing the three bathing tasks of washing, rinsing, and drying, we achieve a fully autonomous cleaning trial on the forearm of a user with severe mobility limitations.}
  \vspace{-15pt}
  \label{fig:wow_figure}
\end{figure}

%% file: src/related-work.tex
\section{Related Work}
In this section, we explore the literature related to robot-assisted bathing and examine research from computer vision and robotics, highlighting works on visual perception and compliant control that have informed the design and development of RABBIT.

\input{figures/src/overview}

\vspace{-5pt}
\subsection{Bathing Assistance Systems}
Robot-assisted bathing encompasses a spectrum of methodologies. Systems such as those in \cite{zlatintsi2017, Dometios2017, zlatintsi2020support} lean towards specialized bathing setups, often necessitating permanent installations. In contrast, our system offers a more accessible solution that avoids the physical strain associated with shower baths and the need for permanent installations. Liu et al. \cite{liu2022characterization} introduces a unique approach to bed bathing with a mesoscale wearable robot that traverses directly on the human body, demonstrating adaptive behaviors and effective cleaning of arms. However, it does not address the practical considerations of employing common bathing agents like soap and water. Amongst bed bathing systems, King et al. \cite{king2015} employ an end-effector-based compliant control emphasizing safety, but the system is largely dependent on human guidance. Shifting the focus to sensing, Erickson et al. \cite{erickson2019multidimensional} propose a six-electrode capacitive sensor for real-time limb contour tracking. While this is a promising development, it emphasizes positional tracking and overlooks the importance of force modulation, which is especially crucial for tasks such as pat drying. Integrating hardware and sensing, Huang et al. \cite{Huang2022} pair an inflatable end-effector with a depth camera, achieving both position and force tracking, thereby addressing one of the limitations of \cite{erickson2019multidimensional}. However, similar to \cite{liu2022characterization, king2015, erickson2019multidimensional}, it does not employ any perception skills to detect transparent, wet, and soapy areas common to bed bathing scenarios. Importantly, the prevailing research also largely misses out on experiments involving individual(s) with mobility limitations, a key demographic for such solutions. RABBIT addresses these limitations by fusing multimodal perception with compliant control for safe and comfortable pHRI, and uniquely demonstrates its efficacy with a person with Multiple Sclerosis (MS).
\vspace{-5pt}
\subsection{Visual Perception in Bed Bathing Tasks}
\label{sec:perception_lit}
Bed bathing introduces unique perception challenges, especially in distinguishing between wet and dry skin. While discerning between frothy soap and water is relatively straightforward due to their distinct visual characteristics, identifying wet skin poses a more complex problem. The subtlety in visual differences between wet and dry skin means that relying solely on RGB features can be insufficient. This challenge aligns with a broader problem in computer vision: transparent object recognition. Although this problem has not been extensively discussed in the context of pHRI, insights from existing work in other areas of robotics provide valuable direction. 
Recognizing transparent objects poses challenges \cite{jiang2023robotic} due to the minimal light absorption or reflection properties of such objects. A popular range of methods, RGB-X fusion, melds RGB images with supplementary sensory input. For instance, in autonomous driving, RGB is combined with depth data to pinpoint reflections from water puddles \cite{kim2016}. However, the reliability of this method diminishes under certain conditions, such as requiring flat reflective surfaces. Spectral techniques, harnessing near-infrared light, detect changes as light intersects with water \cite{mcgunnigle2010}, or identify visual alterations, like surface darkening, induced by water films \cite{Shimano2017}. However, these methods depend on the specific properties of the material and the thickness of the water layer.
In contrast, thermal imaging offers broader applicability. Though transparent in RGB, silicate glass appears colder in thermal imaging due to lack of absorption of long infrared light \cite{Huo_2023} regardless of its thickness. This difference in thermal signatures between glass and its surroundings enables the clear differentiation of transparent glasses from their surroundings. Similarly, the water temperature used for bathing differs significantly from skin temperature, making thermal features useful in our context. Given these advantages, we selected RGB combined with thermal (RGB-T) as the desired sensing modality for RABBIT, enabling the detection of thin-film water while still facilitating the distinction between soap and water.
\vspace{-5pt}
\subsection{Compliant Control in Bed bathing and pHRI}
Bed bathing assistance involves the robot being in consistent and close contact with a user's skin. Excessive force may cause discomfort or even injury, whereas insufficient force could compromise the bathing process. Achieving a balance of force that ensures effective cleaning while maintaining user comfort and safety is imperative. This balance can be achieved through active and passive compliant control mechanisms \cite{DESANTIS2008253, 6630576, doi:10.1089/soro.2013.0002, villani2013null}. Active compliance \cite{komada1988robust, shetty1996active, siciliano1999robot} uses feedback for controlled force exertion, while passive compliance \cite{pratt1995series, beyl2006compliant, ham2009compliant} relies on mechanical design and joint elasticity for adaptability. In the context of bed bathing, King et al. \cite{king2015} and Huang et al. \cite{Huang2022} have looked into implementing compliant controllers using impedance and force control, respectively. Research has looked into dual (software and hardware) compliance for enhanced flexibility. Joonhigh et al. \cite{joonhigh2021variable} present a bubble gripper with inherent passive compliance, modulating the bubble geometry and stiffness to introduce active compliance for grasping a wide range of objects. Additionally, Ashtiani et al. \cite{ashtiani2021hybrid} integrate passive stiffness in a compliant leg joint with a virtual controller for locomotion, addressing sensor noise and sensorimotor delays on uneven terrains and harsh conditions. Given the critical need for safety and adaptability in pHRI, RABBIT adopts dual compliance, aiming for safer and more dependable pHRI.

%% file: figures/src/overview.tex
\begin{figure*}[t!]
  \centering
  {\includegraphics[width=0.99\textwidth]{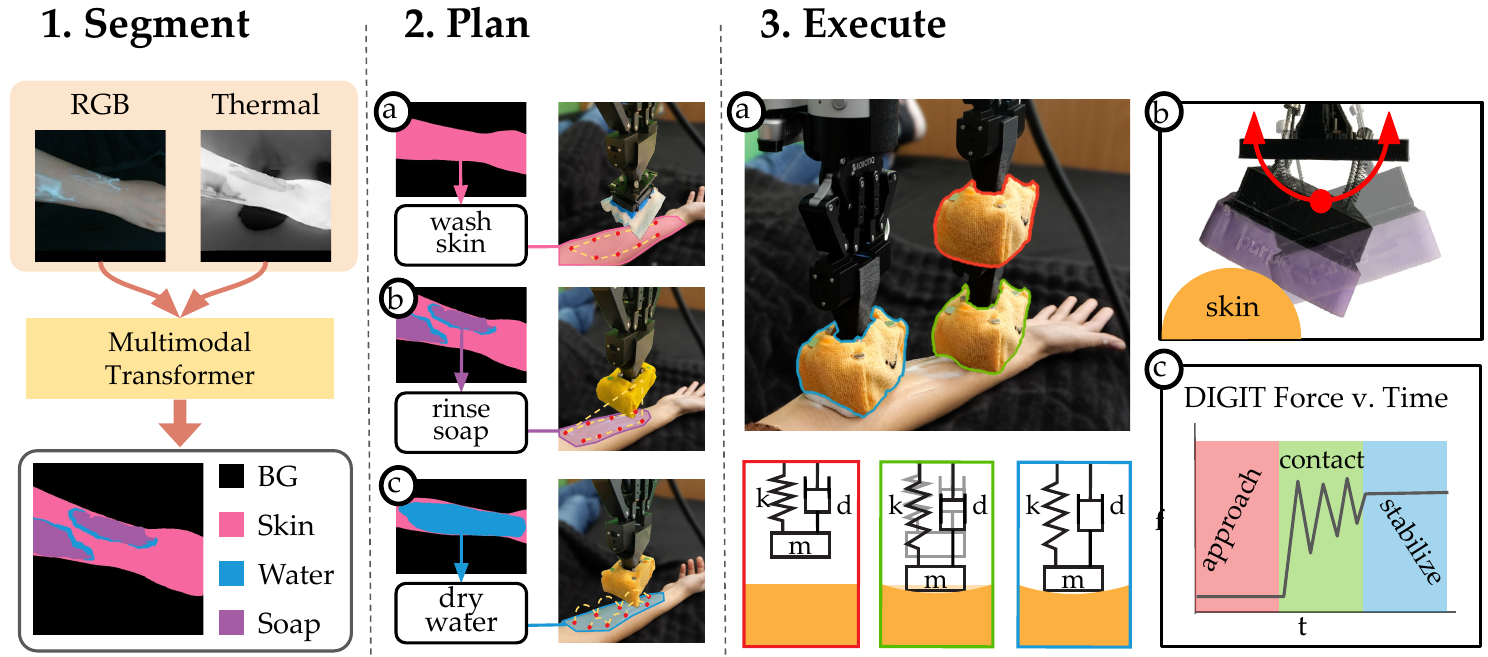}}
  \vspace{-10pt}
  \caption{The key elements of our proposed system. The system first captures a photo of the region it wants to bathe, and segments it as soapy, wet, or dry. Based on the task a path is planned and then the controller executes the plan.}
  \label{fig:overview}
  \vspace{-5pt}
\end{figure*}

%% file: src/methods.tex
\section{Robot-Assisted Bed Bathing}

RABBIT, as depicted in Fig. \ref{fig:overview}, is designed to autonomously execute the three primary tasks of bed bathing: \textbf{washing} (applying soap), \textbf{rinsing} (removing soap with water), and \textbf{drying} (removing the residual water). When a care recipient (CR) lies in a supine position on a hospital bed, RABBIT employs an overhead RGB-D camera to determine the CR's joint skeleton in 3D space (Fig. \ref{fig:experimental_setup}). Leveraging this skeleton pose, the robot's end-effector (EE) positions itself over the body part to be cleaned, such as the right forearm. Upon aligning the EE with the target body part, it captures RGB and thermal images from its onboard cameras. These images are fed into an image segmentation module, which outputs a segmentation mask with four distinct classes: \texttt{background}, \texttt{dry} \texttt{skin}, \texttt{water}, and \texttt{soap}. These masks are pivotal during various stages of the bathing process. For instance, the soap mask identifies the treatment area during rinsing. Depending on the bathing stage, RABBIT identifies the necessary treatment region and communicates it to the planning module. Drawing from expert caregiving guidelines, this module generates a motion plan using predefined motion primitives. This plan is then executed by a task-space compliance controller that employs gain scheduling and impedance control to comfortably maneuver our compliant bathing tool, Scrubby, over the CR's body contours. This section delves into the perception, planning, and control subsystems and how they interact.

\vspace{-5pt}
\subsection{Multimodal Perception for Bed Bathing}
To differentiate between dry, wet, and soapy skin during bed bathing, we cast the problem as image segmentation, aiming to output a mask comprising channels for \texttt{background}, \texttt{dry skin}, \texttt{water}, and \texttt{soap}. Segmentation of skin ensures that we generalize to various arm shapes and contours. Such segmentation is vital for adapting to a variety of arm shapes and contours. Importantly, the fluid nature of wet and soapy areas means that the regions requiring treatment may not align precisely with where water or soap was initially applied. Failing to accurately segment these areas could leave residual soap or water, potentially leading to skin irritation, increased inflammation risk, and enhanced microbial growth. Effective training of deep learning-based image segmentation models for this niche problem requires a domain-specific dataset that captures diverse skin types under various conditions. Thus, our first step was to obtain a feature-rich dataset.

\input{figures/src/experimental_setup}

\textbf{RGB-T Dataset for Bed Bathing}. While existing datasets \cite{saxen2014color, kawulok2014spatial} cater to skin segmentation, none address our unique requirements of samples with soapy and wet skin. Simulations, though useful, can fall short in replicating the thermal properties of real-world interactions, especially between water/soap and human skin. As a result, the need for real-world data is evident in our application. Ensuring dataset diversity in real-world scenarios can be challenging, primarily due to the limited accessibility of varied human data. To address this, we have curated a first-of-its-kind \textit{Synthetic Bathing Perception (SBP)} \textit{dataset} employing both low and high-resolution thermal cameras alongside a Realsense D435i RGB-D camera to capture a wide range of thermal noise characteristics for better generalization. 
The initial phase utilized an MLX90640 thermal camera, capturing images at 24x32 resolution at 16Hz, which yielded 392 RGB-thermal image pairs of the "enhanced" nursing manikin arm and 397 from human participants. This low-resolution camera was chosen to complement the high-resolution P2Pro thermal camera (256x192 resolution at 25Hz), recognizing that different noise characteristics in thermal signatures could enhance the model's ability to generalize to a wider variety of thermal images. The subsequent addition of the high-resolution data added 360 image pairs from the manikin and 240 from human subjects. We created this dataset using the setup outlined in Fig. \ref{fig:experimental_setup}.

\input{figures/src/data_collection}
To ensure skin tone diversity, the manikin arm was painted in six distinct shades based on the Fitzpatrick scale, a standard metric used in dermatology \cite{Ward2017}. After prepping the manikin arm with paint and a sealant for protection and a realistic texture, we heated it to the average human skin temperature of $\sim36.6$\degree C \cite{10.7554/eLife.49555} between samples using an electric heated blanket to ensure realistic thermal signatures that are similar to human skin. This was verified at two positions along the surface of the arm with a point-and-click thermometer, as well as by visual confirmation of even heating using the thermal image stream. For each skin tone, we collected 60 samples: 2 \textit{hair conditions }(True, False) x 5 \textit{categories of coverage} (None, Partial [Soap, Water], Full [Soap, Water]) x 6 \textit{camera poses}. We kept the water temperature fixed at 20$\pm$2\degree C (average room temperature) across all the samples. We verified the water temperature before every water application with a thermometer and applied the water and soap to the arm using our custom wiping tool. This was done to closely emulate the temperatures we may observe during deployment. Despite our best efforts, some thermal image samples exhibit localized artifacts that may appear unrealistic. We included these in the final dataset and collected a smaller dataset with four human participants to mitigate any model-fitting effects. For the images with arm hair, we applied the hair using a small amount of superglue with liquid latex in between for ease of removal. For varied camera perspectives, RGB-D and thermal cameras were affixed to a Kinova Gen3 7DoF arm via a 3D-printed bracket (Fig. \ref{fig:experimental_setup}). Manual annotations were performed using CVAT \cite{cvat}. We provide more details in the supplementary material \cite{rabbit23}.

\textbf{Benchmarking using the SBP dataset}. As discussed in Sec. \ref{sec:perception_lit}, research in image segmentation has increasingly started leveraging RGB and thermal modalities \cite{guo2021, sun2019, ha2017, Fu2022, Zhou2022}. Using the SBP dataset, we benchmarked various state-of-the-art image segmentation techniques, focusing on RGB and RGB-T-based segmentation, aiming to identify the best-performing segmentation model for integration into our system. We evaluated three RGB-T methodologies: MFNet \cite{ha2017}, RTFNet \cite{sun2019}, and CMX \cite{zhang2023cmx}. MFNet and RTFNet are widely known network architectures in the RGB-T image segmentation literature. MFNet is a dual-encoder-based lightweight CNN architecture for image segmentation using RGB-T data that enables real-time inference at $\sim$55FPS. In contrast, RTFNet, an asymmetric encoder-decoder-based architecture, offers 33\% enhanced performance in urban scene segmentation but demonstrates 3x slower inference speed. CMX is a state-of-the-art transformer-based architecture that leverages a Cross-Modal Feature Rectification Module and Feature Fusion Module to rectify and integrate RGB with the thermal input, enhancing image pair alignment and segmentation. Due to the inherent complexity of RGB-only approaches for this task, our benchmark included just one method, SegFormer \cite{xie2021segformer}, which is frequently adopted as a baseline in recent image segmentation research. SegFormer is a transformer-based architecture for image segmentation that is known to effectively capture long-range dependencies, i.e., correspondences between pixels that are further apart in the image. We initialized all methods using ImageNet pretrained weights, followed by fine-tuning.

Our dataset was partitioned into train, validation, and test sets in an 8:1:1 ratio. We employed the Intersection over Union (IoU), which is the overlap between the ground truth and predictions, as our primary evaluation metric. As evidenced in Table \ref{tab:model_performance}, all RGB-T models surpassed SegFormer in terms of skin, water, and mean IoU (mIoU), with SegFormer outperforming RGB-T methods only in background and soap IoU. The superior background performance by SegFormer is likely attributed to the consistent RGB appearance, whereas thermal variations arise from factors like water dripping or heat transfer from the arm. The distinct visual features of soap, such as white and reflective bubbles, could explain its better performance, as the thermal channel may blur the distinction between cool water and soap. A visual comparison of the networks' results is provided in Fig. \ref{fig:model_compare}. Notably, while the RGB-T models showcased similar performance, CMX was adept at capturing finer details like dripping water.

\input{figures/src/model_compare}
\input{src/IOU_table}

\textbf{Sequential Resolution Adaptation (SRA)}. We chose CMX as our primary architecture for its superior segmentation performance and further optimized its performance by exploiting the thermal images collected using thermal sensors of varying resolutions. We adopted a sequential training approach where we initially trained CMX exclusively on low-resolution thermal data from the MLX90640, followed by fine-tuning on high-resolution thermal data from the P2Pro. This strategy aims to utilize the regularization effect of low-resolution data for global feature learning, while the high-resolution data refines the details. Our findings (Table \ref{tab:model_performance}) confirm the efficacy of this training approach, with CMX-SRA exhibiting enhanced IoU performance. We then used this refined model for inference in all our experiments, as discussed in Sec. \ref{sec:user_study}.

\vspace{-5pt}
\subsection{Human Caregiving-inspired Motion Primitives for Bathing}
\label{sec:planning}
Human caregivers take actions during bed bathing that extend beyond merely ensuring adequate coverage and application of soap/water. Expert recommendations inform these practices to enhance the user's health and comfort. We have parameterized our motion primitives to mimic the movements and forces of human caregivers, with validation from experienced OTs. The motion primitives for washing and rinsing replicate the recommended long firm strokes that caregivers use \cite{sorrentino2016mosby} to promote better circulation and venous return. When drying, our system performs the pat drying maneuver \cite{pegram2007clinical, goldenhart2022assisting}, recommended to minimize irritations for patients prone to skin blisters and tears due to prolonged bed and wheelchair usage.

Given these guidelines, we formulate position-and-force parameterized motion primitives. Mathematically, a motion primitive, $P$ is defined as $\{p_0, p_1, p_2,...,p_T\}$, where T is horizon time. \\We define $p_t$ as:
\begin{equation}
p_t = \begin{bmatrix}
       \xi_{d, t}\\
       f_{\text{d}, t}\\
     \end{bmatrix}
\end{equation}

where $\xi_{d, t}$ is the desired 6-DoF pose of the end-effector and $f_{\text{d}, t}$ is the desired 3-DoF force at time $t$. The system identifies the region of interest based on the predicted segmentation mask and the current task: targeting dry mask for washing, soap mask for rinsing, and wet mask for drying. The system then generates a series of 2D waypoints, $W$, as $W = { w_1, w_2, \ldots, w_n }$. It selects each $w_{i}$ based on the horizontal boundaries of the selected region and the planar dimensions of the wiping tool. The system maps these waypoints, drawn over the segmentation mask, to the 3D coordinate space using the corresponding depth image. It then transforms these 3D waypoints into force-position parameterized trajectories by sequentially applying $P_{\text{task}}$ between adjacent waypoints. During the washing phase, $P_{\text{wash}}$ enables the tool to follow a continuous path up and down the arm (Fig. \ref{fig:overview}-2.a) with a fixed desired force. For the rinsing phase, it uses $P_{\text{rinse}}$, making the tool move in straight lines along the length of the forearm in the same direction and repeats this twice to encourage a soap lather and loosen debris on the arm. This design traps the soap along the leading edge of the wiper while the trailing edge compresses against the arm and releases water onto the skin (Fig. \ref{fig:overview}-2.b). Finally, in the drying phase, $P_{\text{dry}}$ makes the tool lift between waypoints and move to the next position before it descends to pat the arm dry along the trajectory (Fig. \ref{fig:overview}-2.c). For each primitive, desired forces and spatial resolutions of the waypoints are tuned, optimizing for efficiency and comfort.

\input{figures/src/system_fig}

\vspace{-5pt}
\subsection{Software and Hardware Compliance}

\input{figures/src/user_study}

\textbf{Gain scheduling Task-Space Compliant Control}. 
We implemented a task-space compliant controller with model-free friction observers to enhance software-side compliance \cite{Khatib1987, Kim2019} using our torque-controlled manipulator. This controller differentiates between the nominal (expected friction-free) motor signal and the measured motor signal to achieve accurate compensation. This allows us to smoothly control the Gen3 arm, which otherwise suffers from poor performance due to unmodeled joint friction and stiction when running in torque mode. We use PID controllers for position and force tracking. The force tracking controller is activated upon contact, taking priority over position tracking along the $z$-axis. Below, we provide the mathematical formulation of our controller.
\begin{align}
    \mathbf{\tau}_{\text{task}} & = -\mathbf{J}^{T}_{\theta} (\mathbf{K}_p \mathbf{e}_x + \mathbf{K}_d \frac{d \mathbf{e}_x}{dt} + \mathbf{K}_i \int \mathbf{e}_x dt + \mathbf{K}_{f,p} \mathbf{e}_f + \mathbf{K}_{f,d} \frac{d \mathbf{e}_f}{dt}) \\
    \mathbf{\tau_{\text{res}}} & =  \mathbf{\tau}_{\text{task}} + \mathbf{\tau}_{\text{fr,nom}} + g_{\theta} 
\end{align}

where $\mathbf{\tau}_{\text{task}}$ is the task-space pose tracking torque, $\mathbf{e}_x$ is the task-space pose tracking error, $\mathbf{e}_f$ is the force tracking error, $\mathbf{K}_p$, $\mathbf{K}_i$, and $\mathbf{K}_d$ are PID gains for the position tracking controller, $\mathbf{K}_{f,p}$ and $\mathbf{K}_{f,d}$ are PD gains for the force tracking controller, $\mathbf{\tau}_{\text{fr,nom}}$ is the nominal friction torque as calculated in \cite{Kim2019}, $g_{\theta}$ is the gravity compensation torque, and $\mathbf{\tau_{\text{res}}}$ is the resultant torque applied to the robot.

To allow for active compliance, our gripper fingers (Robotiq 2F-85) were equipped with a visuo-tactile sensor (DIGIT \cite{Lambeta_2020}). We performed DIGIT sensor calibration using the ATI Nano25 F/T sensor to establish ground truth, employing linear regression to map optical flow to force readings. We use higher stiffness values when performing the tasks of washing and rinsing to ensure thorough cleaning and effective removal of soap. The increased stiffness helps the wiper maintain firm, consistent contact with the surface, enhancing the efficiency of these tasks. Conversely, for actions requiring gentler contact, such as pat drying, we reduce the stiffness to allow for a softer, more delicate touch that is comfortable for the user and prevents skin irritation.

\textbf{Scrubby, a compliant bathing tool}. While software compliance allows for carefully modulating forces for comfortable pHRI, it inherently is restricted by the quality of the force feedback, which may sometimes be noisy and inaccurate for low-cost shear sensing visuo-tactile sensors like DIGIT. We designed a compliant wiper Scrubby (Fig. \ref{fig:overview}) to address this need for residual compliance. Our wiper consists of a top plate with a gripper handle and a bottom plate connected by four low-stiffness springs at the assembly's corners. Tension from a braided wire rope holds the plates parallel when not in contact with a surface. However, the design permits the bottom plate to compress and tilt during contact. While the braided wire columns offer a wide motion range, they also limit excessive shear displacement (Fig. \ref{fig:overview}-2.c). This design focuses on user comfort, with the wiper's compliance allowing it to adapt to the arm's contour without needing a specific contour model.

\vspace{-5pt}
\subsection{System details}
RABBIT uses a Kinova Gen3 7DoF arm in torque-controlled mode. Our system runs on Ubuntu 20.04 Focal and uses ROS noetic to facilitate communication and data visualization. The system incorporates the BehaviorTree library, creating custom nodes to connect various ROS modules. This modular approach is designed for flexibility, allowing components to be upgraded or replaced in line with technological advancements. The system's core operations are facilitated by an AMD Ryzen 9 5900X, operating at 3.7 GHz. This CPU handles data processing and general tasks. For graphics-intensive tasks and deep learning modules like image segmentation, we utilize an NVIDIA GeForce RTX 3090 GPU with 24 GB GDDR6X memory. Additionally, a 32 GB DDR4 RAM supports data flow and caching to reduce latency. In addition, we use a dedicated CPU for running the high-frequency low-level controller to avoid any latency due to the rest of the system.
Fig. \ref{fig:system_fig} shows the flow of our system with the Behavior tree master as the central node that is connected to every other node. At the low level, we switch between gravity compensation, joint-space trajectory control, and task-space compliant control depending on the state of the tree. All the low-level controllers run at 1KHz crucial for stable and responsive compliant control. The pose estimation module operates at a frequency of 12 Hz but is only queried once at the beginning of the treatment. Image segmentation inference coupled with data preprocessing takes $\sim$3 seconds and uses $\sim$10GB of GPU memory. The time delay between capturing the treatment region and starting the treatment is negligible and does not affect the overall system performance. Lastly, our trajectory tracking node that interfaces with the low-level controller works at $\sim$70 Hz, important for smooth tracking of the reference trajectories. These metrics collectively offer insights into RABBIT's system-level operational efficiency.

%% file: figures/src/experimental_setup.tex
\begin{figure}[t!]
      {\includegraphics[width=\columnwidth]{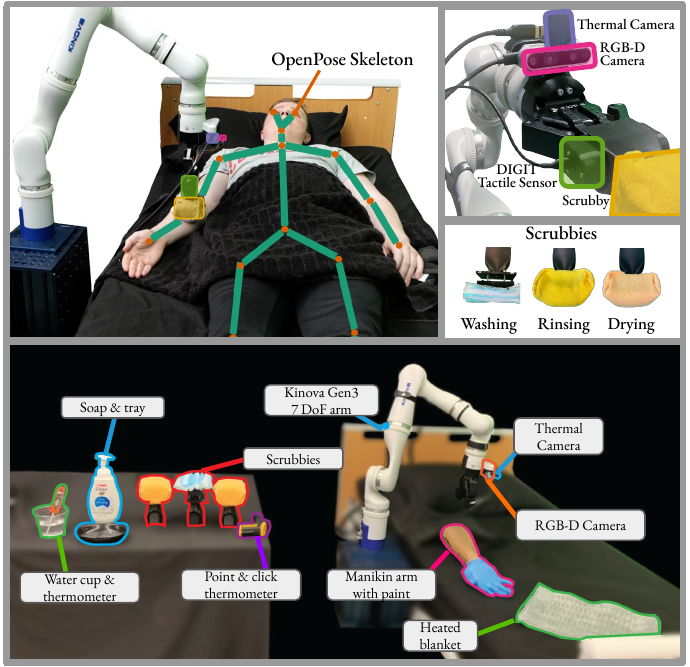}}
      \vspace{-15pt}
  \caption{Top shows user and robot setup, sensors, and bathing tools for each bathing task. The bottom shows the enhanced manikin arm data collection setup.}
  \vspace{-15pt}
  \label{fig:experimental_setup}
\end{figure}

%% file: figures/src/data_collection.tex
\begin{figure}[t!]
      {\includegraphics[width=0.95\columnwidth]{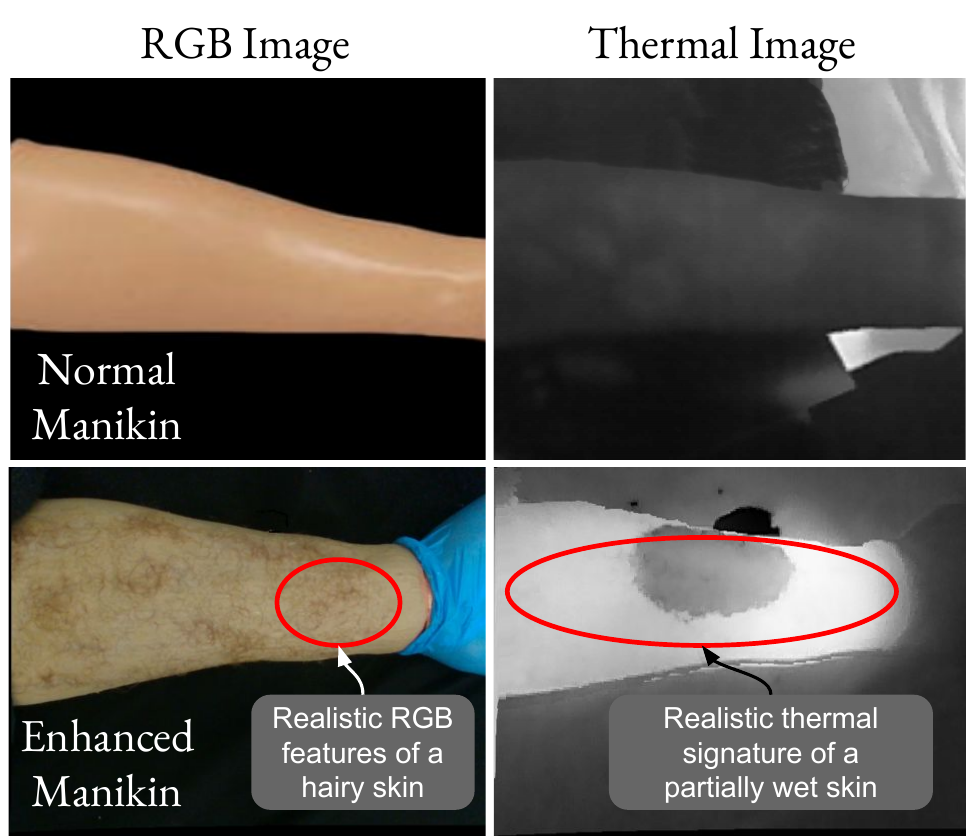}}
      \vspace{-9pt}
  \caption{ \textbf{RGB and thermal images of Enhanced Manikin look realistic and closely resemble features of a human limb.}}
  \vspace{-18pt}
  \label{fig:side by side}
\end{figure}

%% file: figures/src/model_compare.tex
\begin{figure}[t!]
      {\includegraphics[width=\columnwidth]{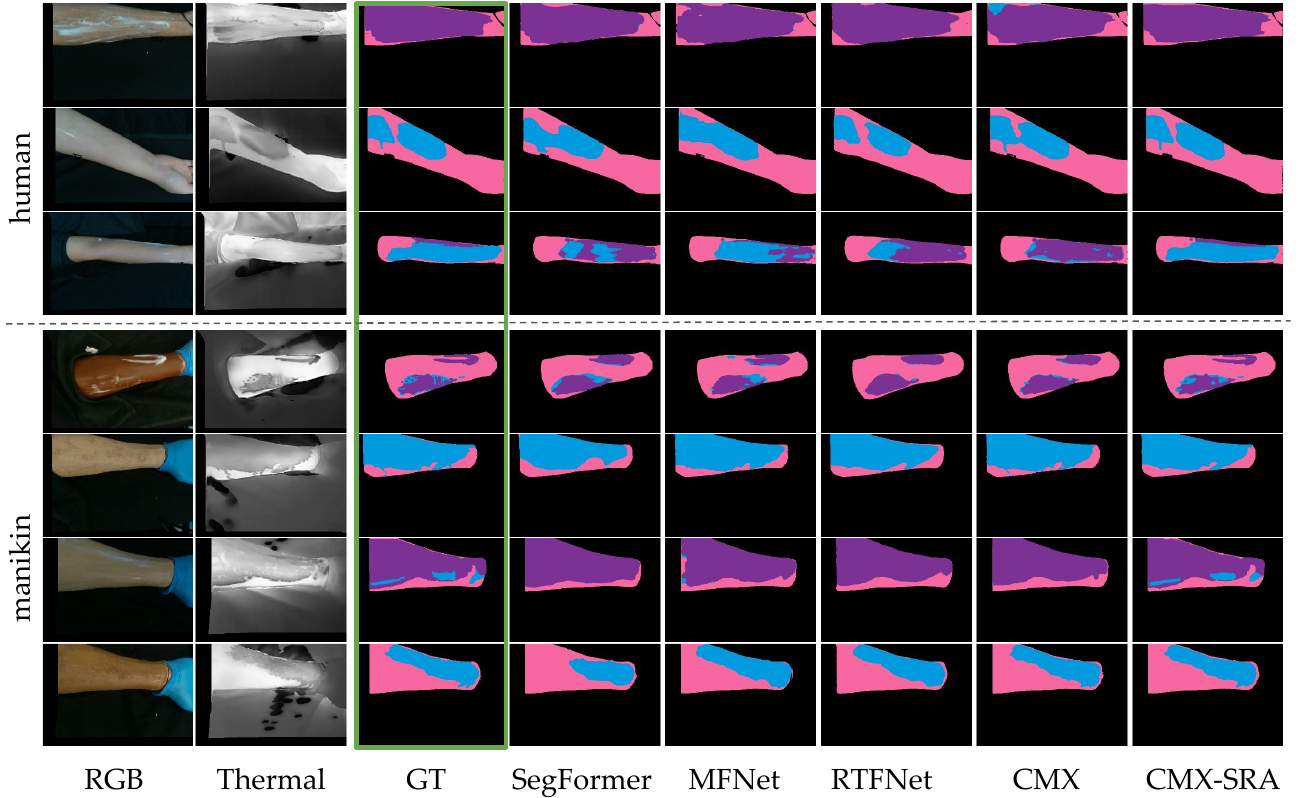}}
      \vspace{-15pt}
  \caption{Qualitative comparison of segmentation models trained on SBP dataset. }
  \vspace{-5pt}
  \label{fig:model_compare}
\end{figure}

%% file: src/IOU_table.tex
\begin{table}[ht!]
\centering
\renewcommand{\arraystretch}{1.2}
\begin{tabularx}{\columnwidth }{>{\centering\arraybackslash\hsize=1.2\hsize\linewidth=\hsize}X >{\centering\arraybackslash\hsize=1.2\hsize\linewidth=\hsize}X >{\centering\arraybackslash\hsize=1.4\hsize\linewidth=\hsize}X >
{\centering\arraybackslash\hsize=0.8\hsize\linewidth=\hsize}X >
{\centering\arraybackslash\hsize=0.8\hsize\linewidth=\hsize}X >
{\centering\arraybackslash\hsize=0.8\hsize\linewidth=\hsize}X >
{\centering\arraybackslash\hsize=0.8\hsize\linewidth=\hsize}X}
\hline
 Model & Type & Background   & Skin & Water & Soap & mIoU \\
\hline
SegFormer & RGB & 99.61 & 84.94 & 75.16 & 87.73 & 86.86 \\
\hline
 RTFNet &  RGB-T & 99.38 & 85.82 &  77.84 & 86.90 & 87.49 \\
\hline
 MFNet & RGB-T & 99.53 & 85.45 & 77.56 & 86.46 & 87.25 \\
\hline
 CMX & RGB-T & 99.61 & 87.75 & 78.55 & 87.03 & 88.23 \\
\hline
 CMX-SRA & RGB-T & 99.72 & $\bf{93.08}$ & $\bf{89.43}$ & \bf$94.93$ & $\bf{94.20}$ \\
\hline

\end{tabularx}
\caption{Comparison of Network classwise and mean IoU (\%) performance on bathing segmentation task}
\label{tab:model_performance}
\vspace{-25pt}
\end{table}

%% file: figures/src/system_fig.tex
\begin{figure}
    \centering
      {\includegraphics[width=\columnwidth]{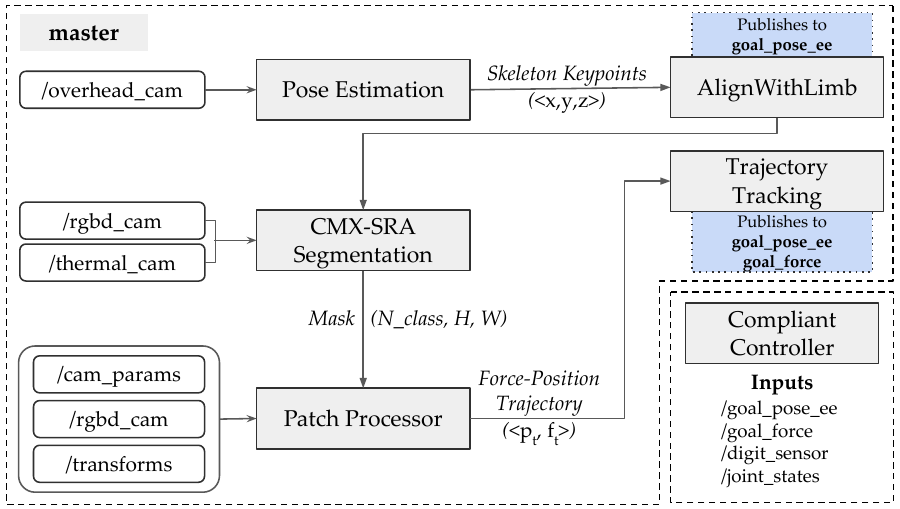}}
      \vspace{-20pt}
  \caption{Our system uses multimodal sensing to sense, plan, and control the robot and perform each of the bathing tasks.}
  \label{fig:system_fig}
  \vspace{-15pt}
\end{figure}

%% file: figures/src/user_study.tex
\begin{figure*}[ht!]
    \centering
      {\includegraphics[width=0.91\textwidth]{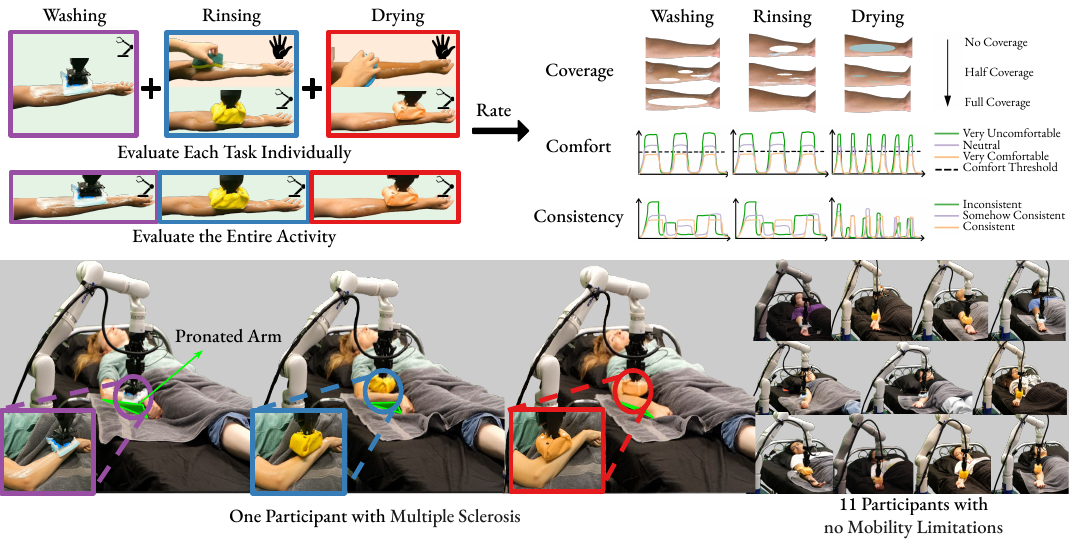}}
  \vspace{-15pt}
  \caption{Study procedure and evaluation criteria for the user studies.}
  \label{fig:user_study_procedure}
  \vspace{-10pt}
\end{figure*}

%% file: src/experiments.tex

%% file: src/user-study.tex
\section{User Study}
\label{sec:user_study}
To evaluate the performance of the RABBIT system in bed bathing tasks, we conducted a user study. This study aimed to assess both individual system components and the performance of the system as a whole. Our study involved 12 participants: 11 individuals without mobility limitations (mean age 24.27 years, SD = 3.38) and a 45-year-old who has lived with Multiple Sclerosis for 26 years. The inclusion of a participant with mobility limitations aimed to validate RABBIT's efficacy for our target user base.
\vspace{-5pt}

\subsection{Study Procedure}
Before initiating the study, we assessed the participants' pre-existing comfort levels with robots: 45.5\% felt comfortable, 45.4\% were neutral, and 9.1\% expressed potential discomfort. After the pre-study questionnaire, we conducted an arm-raising exercise based on OT guidelines. We asked participants to lie down and rest their wrists on our hands. We then lifted their wrists and released them, letting their arms fall naturally onto the bed. Through this exercise, we aimed to help participants experience and replicate the limited arm control that people with mobility limitations often experience. Before the main trials, we also provide a demonstration of the robot's motion for the washing task. This familiarized participants with RABBIT's motions and allowed us to personalize force levels based on their feedback. We perform this personalization as a single-sample procedure where we start with a fixed desired force (as defined in Sec. \ref{sec:planning}) and adjust the values in case the participant finds these values uncomfortable.

\textbf{Individual Task Evaluation}. Before each trial, we prepared participants' forearms accordingly: cleaning and warming them for washing, coating them with soap for rinsing, and moistening them for drying. After completing each task, participants evaluated the robot's performance in terms of coverage, force application, and consistency (Fig. \ref{fig:user_study_procedure}). Regarding coverage, we posed two questions based on: (a) participants' perception of the robot's attempt to cover the target area as a measure of segmentation performance, and (b) participants' perception of the actual area the robot covered as a measure of the planner and tracker performance. For force application, we queried participants on whether they found the overall force applied during the treatment to be comfortable. Regarding consistency, we asked participants to assess how well the robot maintained the same rate of force change for every traversal. We quantify our bathing system's performance with a five-point scale, where 1 indicates the lowest performance and 5 indicates the highest. For instance, in terms of comfort, 1 represents very uncomfortable, 3 represents neutral, and 5 represents very comfortable. We conducted three trials for each of the three bathing tasks.

\textbf{Fully Autonomous Trial Evaluation}. After the individual trials, we conducted a fully autonomous trial where we carried out the washing, rinsing, and drying tasks in sequence. Compared to individual trials, we did not perform any preparation before the treatment since all the tasks were performed by the robot in sequence. Besides the questions asked at the end of each individual trial, we asked two additional questions probing them about the overall coverage performance of the system and perceived comfort for the full trial.

Each participant underwent 10 trials: three tasks multiplied by three individual trials plus one full trial. The study protocol was reviewed and approved by the Institutional Review Board of our organization (IRB0146211). All participants provided written informed consent to participate in the study. We provide other user study-related details in the supplementary material \cite{rabbit23}.

\vspace{-5pt}
\subsection{Results}
For context in interpreting our results, we use the benchmark assumption that expert human caregivers would achieve a perfect rating of 5 in each evaluation category. Overall, our system achieved high mean ratings (see Fig. \ref{fig:analysis}) for all tasks across all evaluation metrics. Furthermore, participants' evaluations for the entire bathing system indicate a level of comfort of 4.83$\pm$0.58 and a performance rating of 4.33$\pm$0.78. Notably, all the participants who initially rated themselves as neutral to uncomfortable about a robot approaching and making contact with them in the pre-study questionnaire later reported feeling comfortable with our system. A one-sample t-test with neutral performance (3) as the hypothesized population mean, yielded statistical significance with $p_{0.005}$ for all the metrics in case of both the individual trials and the full trial.
\input{figures/src/user_study_plots}

The feedback received from participants highlighted the system's capability to provide a comfortable and efficient bed bath. As one participant described, the consistency in pressure application was "like that of a caregiver." Another remarked on the system's systematic approach during washing, noting that "The pressure on soap application was well done." Several participants found the system's predefined speed and force appropriate and comforting. Feedback such as "Robot was gentle and did not move very fast" indicated the balance our system strikes between efficiency and user comfort. A noteworthy comment from one participant was that the system seemed to "care" and took its
time to complete tasks accurately.

Our qualitative observations with the MS participant emphasized the adaptability of our system. Due to their medical condition, they naturally assumed a pronated arm position (see Fig. \ref{fig:user_study_procedure}). Although we had not explicitly calibrated our system for this posture, it successfully managed all three tasks for this participant. This interaction highlights the potential adaptability of our system to a range of user conditions. They mentioned that their most significant source of discomfort during the study was the soap and water left underneath the arm, underscoring the need for the system to reach and clean hidden limb regions effectively. When asked about improvements, they suggested the system should target harder-to-reach areas like finger and toe crevices, pointing to an essential area for future enhancement.

%% file: figures/src/user_study_plots.tex
\begin{figure}[t!]
      {\includegraphics[width=0.91\columnwidth]{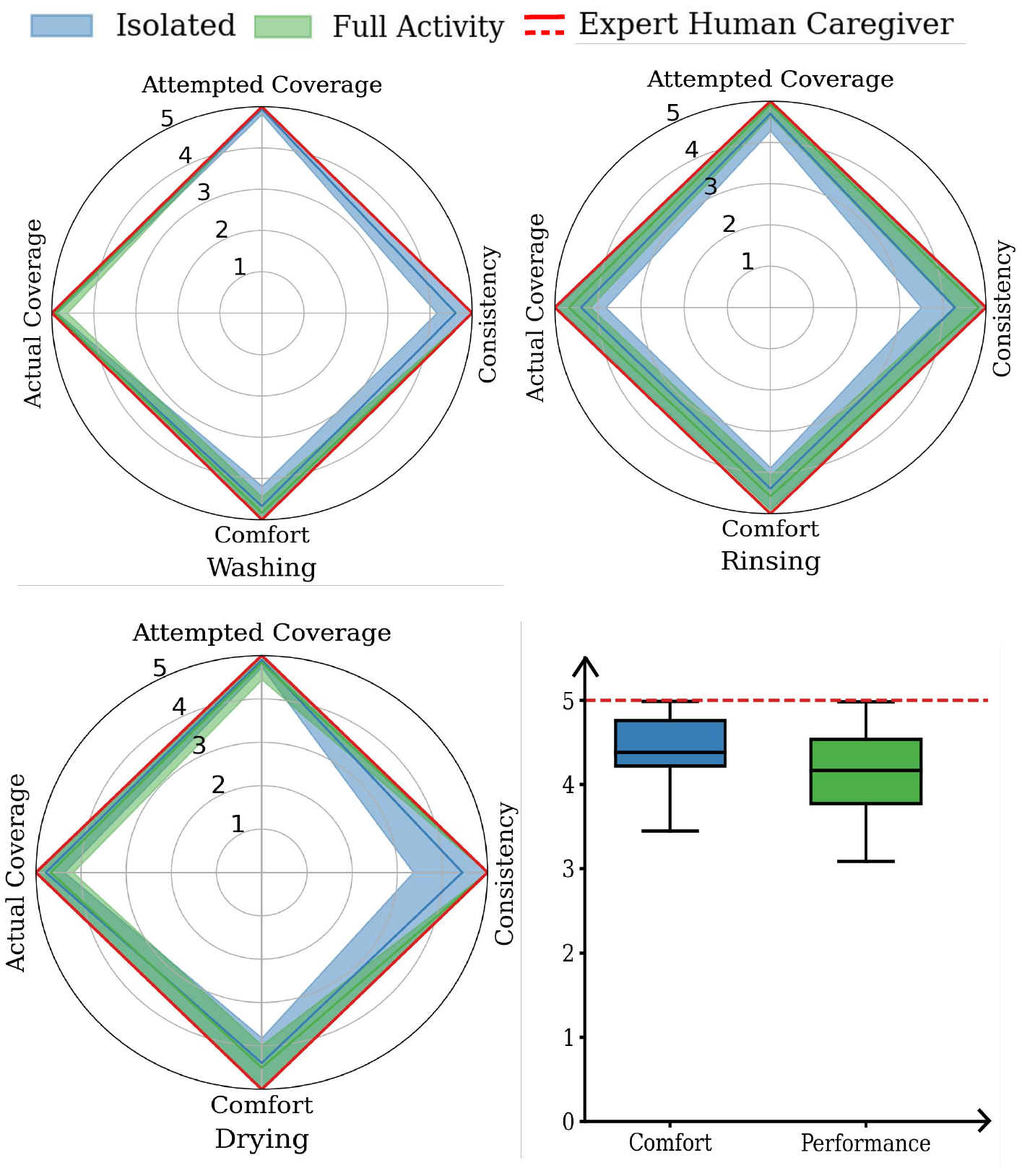}}
      \vspace{-13pt}
  \caption{Participants in the user study gave high ratings across all four metrics for each task, for both isolated trials and full activity.}
  \label{fig:analysis}
  \vspace{-10pt}
\end{figure}

%% file: src/discussion.tex
\section{Discussion}
To the best of our knowledge, RABBIT is the first autonomous robot-assisted bed bathing system that performs the three primary bathing tasks of washing, rinsing, and drying. Although our preliminary study involved only one participant with MS, it suggests the potential of our system to assist care recipients with severe mobility limitations. However, developing a system that offers meaningful, long-term assistance entails addressing several challenges, particularly the need for comprehensive coverage and sensitivity to individual needs. Enhancements are necessary to enable full-body bathing, which involves reaching various body parts and treating them effectively. This includes integrating navigation capabilities to reach accessible body parts, limb manipulation to reach hidden regions, and dexterous manipulation for sensitive and intricate areas like the face and finger crevices (regions suggested by the MS participant). Additionally, our current open-loop planning architecture needs improvements to serve a diverse population better. The problem of treating residual soap and water effectively, along with adapting to involuntary movements like tremors, necessitates incorporating reactive planning strategies and adopting closed-loop visuo-tactile motor policies. An essential aspect of physical human-robot interaction (pHRI) is integrating user preferences and feedback into the robotic system. While our current system performs an initial calibration, it may be limiting for cases where the user might change preferences during the treatment. Future work can focus on online adaptation to user preferences, particularly in determining where users are comfortable with the robot making contact and the amount of force they find acceptable. This requires a delicate balance between accommodating individual comfort and ensuring effective task performance. As we progress, RABBIT can be improved to make it more responsive, attuned to users' needs, and focused on individual preferences.

%% file: main.bbl

\begin{thebibliography}{57}


\ifx \showCODEN    \undefined \def \showCODEN     #1{\unskip}     \fi
\ifx \showDOI      \undefined \def \showDOI       #1{#1}\fi
\ifx \showISBNx    \undefined \def \showISBNx     #1{\unskip}     \fi
\ifx \showISBNxiii \undefined \def \showISBNxiii  #1{\unskip}     \fi
\ifx \showISSN     \undefined \def \showISSN      #1{\unskip}     \fi
\ifx \showLCCN     \undefined \def \showLCCN      #1{\unskip}     \fi
\ifx \shownote     \undefined \def \shownote      #1{#1}          \fi
\ifx \showarticletitle \undefined \def \showarticletitle #1{#1}   \fi
\ifx \showURL      \undefined \def \showURL       {\relax}        \fi
\providecommand\bibfield[2]{#2}
\providecommand\bibinfo[2]{#2}
\providecommand\natexlab[1]{#1}
\providecommand\showeprint[2][]{arXiv:#2}

\bibitem[Abdi et~al\mbox{.}(2021)]%
        {abdi2021emerging}
\bibfield{author}{\bibinfo{person}{Sarah Abdi}, \bibinfo{person}{Irene Kitsara}, \bibinfo{person}{Mark~S Hawley}, {and} \bibinfo{person}{LP de Witte}.} \bibinfo{year}{2021}\natexlab{}.
\newblock \showarticletitle{Emerging technologies and their potential for generating new assistive technologies}.
\newblock \bibinfo{journal}{\emph{Assistive Technology}} \bibinfo{volume}{33}, \bibinfo{number}{sup1} (\bibinfo{year}{2021}), \bibinfo{pages}{17--26}.
\newblock


\bibitem[Alberta(2024)]%
        {albertacare}
\bibfield{author}{\bibinfo{person}{MyHealth Alberta}.} \bibinfo{year}{2024}\natexlab{}.
\newblock \bibinfo{title}{Caregiving: How to Give a Bed Bath}.
\newblock
\newblock
\urldef\tempurl%
\url{https://myhealth.alberta.ca/Health/Pages/conditions.aspx?hwid=abp9705&lang=en-ca}
\showURL{%
\tempurl}
\newblock
\shownote{(Accessed: 3rd January, 2024)}.


\bibitem[Ashtiani et~al\mbox{.}(2021)]%
        {ashtiani2021hybrid}
\bibfield{author}{\bibinfo{person}{Milad~Shafiee Ashtiani}, \bibinfo{person}{Alborz Aghamaleki~Sarvestani}, {and} \bibinfo{person}{Alexander Badri-Spr{\"o}witz}.} \bibinfo{year}{2021}\natexlab{}.
\newblock \showarticletitle{Hybrid parallel compliance allows robots to operate with sensorimotor delays and low control frequencies}.
\newblock \bibinfo{journal}{\emph{Frontiers in Robotics and AI}}  \bibinfo{volume}{8} (\bibinfo{year}{2021}), \bibinfo{pages}{645748}.
\newblock


\bibitem[Beyl et~al\mbox{.}(2006)]%
        {beyl2006compliant}
\bibfield{author}{\bibinfo{person}{Pieter Beyl}, \bibinfo{person}{Bram Vanderborght}, \bibinfo{person}{Ronald Van~Ham}, \bibinfo{person}{Micha{\"e}l Van~Damme}, \bibinfo{person}{Rino Versluys}, {and} \bibinfo{person}{Dirk Lefeber}.} \bibinfo{year}{2006}\natexlab{}.
\newblock \showarticletitle{Compliant actuation in new robotic applications}. In \bibinfo{booktitle}{\emph{NCTAM06--7th National Congress on Theoretical and Applied Mechanics}}.
\newblock


\bibitem[CareChannel(2019)]%
        {CareChannel_2019}
\bibfield{author}{\bibinfo{person}{CareChannel}.} \bibinfo{year}{2019}\natexlab{}.
\newblock \bibinfo{title}{How to give a Bed Bath in the Home - Tips for Caregivers}.
\newblock
\newblock
\urldef\tempurl%
\url{https://www.youtube.com/watch?v=HRfFdgch968}
\showURL{%
\tempurl}


\bibitem[Czuba et~al\mbox{.}(2012)]%
        {czuba2012ergonomic}
\bibfield{author}{\bibinfo{person}{Laura~Rae Czuba}, \bibinfo{person}{Carolyn~M Sommerich}, {and} \bibinfo{person}{Steven~A Lavender}.} \bibinfo{year}{2012}\natexlab{}.
\newblock \showarticletitle{Ergonomic and safety risk factors in home health care: Exploration and assessment of alternative interventions}.
\newblock \bibinfo{journal}{\emph{Work}} \bibinfo{volume}{42}, \bibinfo{number}{3} (\bibinfo{year}{2012}), \bibinfo{pages}{341--353}.
\newblock


\bibitem[Darragh et~al\mbox{.}(2015)]%
        {darragh2015musculoskeletal}
\bibfield{author}{\bibinfo{person}{Amy~R Darragh}, \bibinfo{person}{Carolyn~M Sommerich}, \bibinfo{person}{Steven~A Lavender}, \bibinfo{person}{Kelly~J Tanner}, \bibinfo{person}{Kasey Vogel}, {and} \bibinfo{person}{Marc Campo}.} \bibinfo{year}{2015}\natexlab{}.
\newblock \showarticletitle{Musculoskeletal discomfort, physical demand, and caregiving activities in informal caregivers}.
\newblock \bibinfo{journal}{\emph{Journal of Applied Gerontology}} \bibinfo{volume}{34}, \bibinfo{number}{6} (\bibinfo{year}{2015}), \bibinfo{pages}{734--760}.
\newblock


\bibitem[{De Santis} et~al\mbox{.}(2008)]%
        {DESANTIS2008253}
\bibfield{author}{\bibinfo{person}{Agostino {De Santis}}, \bibinfo{person}{Bruno Siciliano}, \bibinfo{person}{Alessandro {De Luca}}, {and} \bibinfo{person}{Antonio Bicchi}.} \bibinfo{year}{2008}\natexlab{}.
\newblock \showarticletitle{An atlas of physical human--robot interaction}.
\newblock \bibinfo{journal}{\emph{Mechanism and Machine Theory}} \bibinfo{volume}{43}, \bibinfo{number}{3} (\bibinfo{year}{2008}), \bibinfo{pages}{253--270}.
\newblock
\showISSN{0094-114X}
\urldef\tempurl%
\url{https://doi.org/10.1016/j.mechmachtheory.2007.03.003}
\showDOI{\tempurl}


\bibitem[Dometios et~al\mbox{.}(2017)]%
        {Dometios2017}
\bibfield{author}{\bibinfo{person}{Athanasios~C. Dometios}, \bibinfo{person}{Xanthi~S. Papageorgiou}, \bibinfo{person}{Antonis Arvanitakis}, \bibinfo{person}{Costas~S. Tzafestas}, {and} \bibinfo{person}{Petros Maragos}.} \bibinfo{year}{2017}\natexlab{}.
\newblock \showarticletitle{Real-time end-effector motion behavior planning approach using on-line point-cloud data towards a user adaptive assistive bath robot}. In \bibinfo{booktitle}{\emph{2017 IEEE/RSJ International Conference on Intelligent Robots and Systems (IROS)}}. \bibinfo{pages}{5031--5036}.
\newblock
\urldef\tempurl%
\url{https://doi.org/10.1109/IROS.2017.8206387}
\showDOI{\tempurl}


\bibitem[Downey and Lloyd(2008)]%
        {downey2008bed}
\bibfield{author}{\bibinfo{person}{Lindsey Downey} {and} \bibinfo{person}{Hilary Lloyd}.} \bibinfo{year}{2008}\natexlab{}.
\newblock \showarticletitle{Bed bathing patients in hospital}.
\newblock \bibinfo{journal}{\emph{Nursing standard}} \bibinfo{volume}{22}, \bibinfo{number}{34} (\bibinfo{year}{2008}), \bibinfo{pages}{35--40}.
\newblock


\bibitem[Dunlop et~al\mbox{.}(1997)]%
        {dunlop1997disability}
\bibfield{author}{\bibinfo{person}{Dorothy~D Dunlop}, \bibinfo{person}{Susan~L Hughes}, {and} \bibinfo{person}{Larry~M Manheim}.} \bibinfo{year}{1997}\natexlab{}.
\newblock \showarticletitle{Disability in activities of daily living: patterns of change and a hierarchy of disability.}
\newblock \bibinfo{journal}{\emph{American journal of public health}} \bibinfo{volume}{87}, \bibinfo{number}{3} (\bibinfo{year}{1997}), \bibinfo{pages}{378--383}.
\newblock


\bibitem[Erickson et~al\mbox{.}(2019)]%
        {erickson2019multidimensional}
\bibfield{author}{\bibinfo{person}{Zackory Erickson}, \bibinfo{person}{Henry~M Clever}, \bibinfo{person}{Vamsee Gangaram}, \bibinfo{person}{Greg Turk}, \bibinfo{person}{C~Karen Liu}, {and} \bibinfo{person}{Charles~C Kemp}.} \bibinfo{year}{2019}\natexlab{}.
\newblock \showarticletitle{Multidimensional capacitive sensing for robot-assisted dressing and bathing}. In \bibinfo{booktitle}{\emph{2019 IEEE 16th International Conference on Rehabilitation Robotics (ICORR)}}. IEEE, \bibinfo{pages}{224--231}.
\newblock


\bibitem[Fu et~al\mbox{.}(2022)]%
        {Fu2022}
\bibfield{author}{\bibinfo{person}{Yanping Fu}, \bibinfo{person}{Qiaoqiao Chen}, {and} \bibinfo{person}{Haifeng Zhao}.} \bibinfo{year}{2022}\natexlab{}.
\newblock \showarticletitle{CGFNet: cross-guided fusion network for RGB-thermal semantic segmentation}.
\newblock \bibinfo{journal}{\emph{The Visual Computer}}  \bibinfo{volume}{38} (\bibinfo{year}{2022}).
\newblock
Issue 9.
\urldef\tempurl%
\url{https://doi.org/10.1007/s00371-022-02559-2}
\showDOI{\tempurl}


\bibitem[Goldenhart and Nagy(2022)]%
        {goldenhart2022assisting}
\bibfield{author}{\bibinfo{person}{Alyssa~L Goldenhart} {and} \bibinfo{person}{Hassan Nagy}.} \bibinfo{year}{2022}\natexlab{}.
\newblock \showarticletitle{Assisting patients with personal hygiene}.
\newblock In \bibinfo{booktitle}{\emph{StatPearls [Internet]}}. \bibinfo{publisher}{Statpearls Publishing}.
\newblock


\bibitem[Guo et~al\mbox{.}(2021)]%
        {guo2021}
\bibfield{author}{\bibinfo{person}{Zhifeng Guo}, \bibinfo{person}{Xu Li}, \bibinfo{person}{Qimin Xu}, {and} \bibinfo{person}{Zhengliang Sun}.} \bibinfo{year}{2021}\natexlab{}.
\newblock \showarticletitle{Robust semantic segmentation based on RGB-thermal in variable lighting scenes}.
\newblock \bibinfo{journal}{\emph{Measurement}}  \bibinfo{volume}{186} (\bibinfo{year}{2021}), \bibinfo{pages}{110176}.
\newblock
\showISSN{0263-2241}
\urldef\tempurl%
\url{https://doi.org/10.1016/j.measurement.2021.110176}
\showDOI{\tempurl}


\bibitem[Ha et~al\mbox{.}(2017)]%
        {ha2017}
\bibfield{author}{\bibinfo{person}{Qishen Ha}, \bibinfo{person}{Kohei Watanabe}, \bibinfo{person}{Takumi Karasawa}, \bibinfo{person}{Yoshitaka Ushiku}, {and} \bibinfo{person}{Tatsuya Harada}.} \bibinfo{year}{2017}\natexlab{}.
\newblock \showarticletitle{MFNet: Towards real-time semantic segmentation for autonomous vehicles with multi-spectral scenes}. In \bibinfo{booktitle}{\emph{2017 IEEE/RSJ International Conference on Intelligent Robots and Systems (IROS)}}. \bibinfo{pages}{5108--5115}.
\newblock
\urldef\tempurl%
\url{https://doi.org/10.1109/IROS.2017.ha2017}
\showDOI{\tempurl}


\bibitem[Ham et~al\mbox{.}(2009)]%
        {ham2009compliant}
\bibfield{author}{\bibinfo{person}{Ronald Ham}, \bibinfo{person}{Thomas Sugar}, \bibinfo{person}{Bram Vanderborght}, \bibinfo{person}{Kevin Hollander}, {and} \bibinfo{person}{Dirk Lefeber}.} \bibinfo{year}{2009}\natexlab{}.
\newblock \showarticletitle{Compliant actuator designs}.
\newblock \bibinfo{journal}{\emph{IEEE Robotics \& Automation Magazine}} \bibinfo{volume}{3}, \bibinfo{number}{16} (\bibinfo{year}{2009}), \bibinfo{pages}{81--94}.
\newblock


\bibitem[Huang et~al\mbox{.}(2022)]%
        {Huang2022}
\bibfield{author}{\bibinfo{person}{Isabella Huang}, \bibinfo{person}{Dylan Chow}, {and} \bibinfo{person}{Ruzena Bajcsy}.} \bibinfo{year}{2022}\natexlab{}.
\newblock \showarticletitle{Soft Tactile Contour Following for Robot-Assisted Wiping and Bathing}. In \bibinfo{booktitle}{\emph{2022 IEEE/RSJ International Conference on Intelligent Robots and Systems (IROS)}}. \bibinfo{pages}{7797--7802}.
\newblock
\urldef\tempurl%
\url{https://doi.org/10.1109/IROS47612.2022.9982071}
\showDOI{\tempurl}


\bibitem[Huo et~al\mbox{.}(2023)]%
        {Huo_2023}
\bibfield{author}{\bibinfo{person}{Dong Huo}, \bibinfo{person}{Jian Wang}, \bibinfo{person}{Yiming Qian}, {and} \bibinfo{person}{Yee-Hong Yang}.} \bibinfo{year}{2023}\natexlab{}.
\newblock \showarticletitle{Glass Segmentation With {RGB}-Thermal Image Pairs}.
\newblock \bibinfo{journal}{\emph{{IEEE} Transactions on Image Processing}}  \bibinfo{volume}{32} (\bibinfo{year}{2023}), \bibinfo{pages}{1911--1926}.
\newblock
\urldef\tempurl%
\url{https://doi.org/10.1109/tip.2023.3256762}
\showDOI{\tempurl}


\bibitem[Jakopin et~al\mbox{.}(2017)]%
        {jakopin2017unobtrusive}
\bibfield{author}{\bibinfo{person}{Bla{\v{z}} Jakopin}, \bibinfo{person}{Matja{\v{z}} Mihelj}, {and} \bibinfo{person}{Marko Munih}.} \bibinfo{year}{2017}\natexlab{}.
\newblock \showarticletitle{An unobtrusive measurement method for assessing physiological response in physical human--robot interaction}.
\newblock \bibinfo{journal}{\emph{IEEE Transactions on Human-Machine Systems}} \bibinfo{volume}{47}, \bibinfo{number}{4} (\bibinfo{year}{2017}), \bibinfo{pages}{474--485}.
\newblock


\bibitem[Jiang et~al\mbox{.}(2023)]%
        {jiang2023robotic}
\bibfield{author}{\bibinfo{person}{Jiaqi Jiang}, \bibinfo{person}{Guanqun Cao}, \bibinfo{person}{Jiankang Deng}, \bibinfo{person}{Thanh-Toan Do}, {and} \bibinfo{person}{Shan Luo}.} \bibinfo{year}{2023}\natexlab{}.
\newblock \showarticletitle{Robotic Perception of Transparent Objects: A Review}.
\newblock \bibinfo{journal}{\emph{arXiv preprint arXiv:2304.00157}} (\bibinfo{year}{2023}).
\newblock


\bibitem[Joonhigh et~al\mbox{.}(2021)]%
        {joonhigh2021variable}
\bibfield{author}{\bibinfo{person}{Sihah Joonhigh}, \bibinfo{person}{Naveen Kuppuswamy}, \bibinfo{person}{Andrew Beaulieu}, \bibinfo{person}{Alex Alspach}, {and} \bibinfo{person}{Russ Tedrake}.} \bibinfo{year}{2021}\natexlab{}.
\newblock \showarticletitle{Variable compliance and geometry regulation of Soft-Bubble grippers with active pressure control}. In \bibinfo{booktitle}{\emph{2021 IEEE 4th International Conference on Soft Robotics (RoboSoft)}}. IEEE, \bibinfo{pages}{169--175}.
\newblock


\bibitem[Kawulok et~al\mbox{.}(2014)]%
        {kawulok2014spatial}
\bibfield{author}{\bibinfo{person}{Michal Kawulok}, \bibinfo{person}{Jolanta Kawulok}, {and} \bibinfo{person}{Jakub Nalepa}.} \bibinfo{year}{2014}\natexlab{}.
\newblock \showarticletitle{Spatial-based skin detection using discriminative skin-presence features}.
\newblock \bibinfo{journal}{\emph{Pattern Recognition Letters}}  \bibinfo{volume}{41} (\bibinfo{year}{2014}), \bibinfo{pages}{3--13}.
\newblock


\bibitem[Khatib(1987)]%
        {Khatib1987}
\bibfield{author}{\bibinfo{person}{O. Khatib}.} \bibinfo{year}{1987}\natexlab{}.
\newblock \showarticletitle{A unified approach for motion and force control of robot manipulators: The operational space formulation}.
\newblock \bibinfo{journal}{\emph{IEEE Journal on Robotics and Automation}} \bibinfo{volume}{3}, \bibinfo{number}{1} (\bibinfo{year}{1987}), \bibinfo{pages}{43--53}.
\newblock
\urldef\tempurl%
\url{https://doi.org/10.1109/JRA.1987.1087068}
\showDOI{\tempurl}


\bibitem[Kim et~al\mbox{.}(2016)]%
        {kim2016}
\bibfield{author}{\bibinfo{person}{Jisu Kim}, \bibinfo{person}{Jeonghyun Baek}, \bibinfo{person}{Hyukdoo Choi}, {and} \bibinfo{person}{Euntai Kim}.} \bibinfo{year}{2016}\natexlab{}.
\newblock \showarticletitle{Wet area and puddle detection for Advanced Driver Assistance Systems (ADAS) using a stereo camera}.
\newblock \bibinfo{journal}{\emph{International Journal of Control, Automation and Systems}}  \bibinfo{volume}{14} (\bibinfo{year}{2016}), \bibinfo{pages}{263--271}.
\newblock


\bibitem[Kim et~al\mbox{.}(2019)]%
        {Kim2019}
\bibfield{author}{\bibinfo{person}{Min~Jun Kim}, \bibinfo{person}{Fabian Beck}, \bibinfo{person}{Christian Ott}, {and} \bibinfo{person}{Alin Albu-Schäffer}.} \bibinfo{year}{2019}\natexlab{}.
\newblock \showarticletitle{Model-Free Friction Observers for Flexible Joint Robots With Torque Measurements}.
\newblock \bibinfo{journal}{\emph{IEEE Transactions on Robotics}} \bibinfo{volume}{35}, \bibinfo{number}{6} (\bibinfo{year}{2019}), \bibinfo{pages}{1508--1515}.
\newblock
\urldef\tempurl%
\url{https://doi.org/10.1109/TRO.2019.2926496}
\showDOI{\tempurl}


\bibitem[King et~al\mbox{.}(2010)]%
        {king2015}
\bibfield{author}{\bibinfo{person}{Chih-Hung King}, \bibinfo{person}{Tiffany~L. Chen}, \bibinfo{person}{Advait Jain}, {and} \bibinfo{person}{Charles~C. Kemp}.} \bibinfo{year}{2010}\natexlab{}.
\newblock \showarticletitle{Towards an assistive robot that autonomously performs bed baths for patient hygiene}. In \bibinfo{booktitle}{\emph{2010 IEEE/RSJ International Conference on Intelligent Robots and Systems}}. \bibinfo{pages}{319--324}.
\newblock
\urldef\tempurl%
\url{https://doi.org/10.1109/IROS.2010.5649101}
\showDOI{\tempurl}


\bibitem[Komada and Ohnishi(1988)]%
        {komada1988robust}
\bibfield{author}{\bibinfo{person}{S Komada} {and} \bibinfo{person}{K Ohnishi}.} \bibinfo{year}{1988}\natexlab{}.
\newblock \showarticletitle{Robust force and compliance control of robotic manipulator}. In \bibinfo{booktitle}{\emph{Proceedings. 14 Annual Conference of Industrial Electronics Society}}, Vol.~\bibinfo{volume}{1}. IEEE, \bibinfo{pages}{20--25}.
\newblock


\bibitem[Lab(2024)]%
        {rabbit23}
\bibfield{author}{\bibinfo{person}{EmPRISE Lab}.} \bibinfo{year}{2024}\natexlab{}.
\newblock \bibinfo{title}{{RABBIT} website}.
\newblock
\newblock
\urldef\tempurl%
\url{https://emprise.cs.cornell.edu/rabbit}
\showURL{%
\tempurl}
\newblock
\shownote{(Accessed: 3rd January, 2024)}.


\bibitem[Lambeta et~al\mbox{.}(2020)]%
        {Lambeta_2020}
\bibfield{author}{\bibinfo{person}{Mike Lambeta}, \bibinfo{person}{Po-Wei Chou}, \bibinfo{person}{Stephen Tian}, \bibinfo{person}{Brian Yang}, \bibinfo{person}{Benjamin Maloon}, \bibinfo{person}{Victoria~Rose Most}, \bibinfo{person}{Dave Stroud}, \bibinfo{person}{Raymond Santos}, \bibinfo{person}{Ahmad Byagowi}, \bibinfo{person}{Gregg Kammerer}, \bibinfo{person}{Dinesh Jayaraman}, {and} \bibinfo{person}{Roberto Calandra}.} \bibinfo{year}{2020}\natexlab{}.
\newblock \showarticletitle{{DIGIT}: A Novel Design for a Low-Cost Compact High-Resolution Tactile Sensor With Application to In-Hand Manipulation}.
\newblock \bibinfo{journal}{\emph{{IEEE} Robotics and Automation Letters}} \bibinfo{volume}{5}, \bibinfo{number}{3} (\bibinfo{date}{jul} \bibinfo{year}{2020}), \bibinfo{pages}{3838--3845}.
\newblock
\urldef\tempurl%
\url{https://doi.org/10.1109/lra.2020.2977257}
\showDOI{\tempurl}


\bibitem[Liu et~al\mbox{.}(2022b)]%
        {liu2022characterization}
\bibfield{author}{\bibinfo{person}{Fukang Liu}, \bibinfo{person}{Vaidehi Patil}, \bibinfo{person}{Zackory Erickson}, {and} \bibinfo{person}{Zeynep Temel}.} \bibinfo{year}{2022}\natexlab{b}.
\newblock \showarticletitle{Characterization of a Meso-Scale Wearable Robot for Bathing Assistance}. In \bibinfo{booktitle}{\emph{2022 IEEE International Conference on Robotics and Biomimetics (ROBIO)}}. IEEE, \bibinfo{pages}{2146--2152}.
\newblock


\bibitem[Liu et~al\mbox{.}(2020)]%
        {liu2020simultaneously}
\bibfield{author}{\bibinfo{person}{Shuangjun Liu}, \bibinfo{person}{Xiaofei Huang}, \bibinfo{person}{Nihang Fu}, \bibinfo{person}{Cheng Li}, \bibinfo{person}{Zhongnan Su}, {and} \bibinfo{person}{Sarah Ostadabbas}.} \bibinfo{year}{2020}\natexlab{}.
\newblock \showarticletitle{Simultaneously-collected multimodal lying pose dataset: Towards in-bed human pose monitoring under adverse vision conditions}.
\newblock \bibinfo{journal}{\emph{arXiv preprint arXiv:2008.08735}} (\bibinfo{year}{2020}).
\newblock


\bibitem[Liu et~al\mbox{.}(2022a)]%
        {liu2022simultaneously}
\bibfield{author}{\bibinfo{person}{Shuangjun Liu}, \bibinfo{person}{Xiaofei Huang}, \bibinfo{person}{Nihang Fu}, \bibinfo{person}{Cheng Li}, \bibinfo{person}{Zhongnan Su}, {and} \bibinfo{person}{Sarah Ostadabbas}.} \bibinfo{year}{2022}\natexlab{a}.
\newblock \showarticletitle{Simultaneously-collected multimodal lying pose dataset: Enabling in-bed human pose monitoring}.
\newblock \bibinfo{journal}{\emph{IEEE Transactions on Pattern Analysis and Machine Intelligence}} \bibinfo{volume}{45}, \bibinfo{number}{1} (\bibinfo{year}{2022}), \bibinfo{pages}{1106--1118}.
\newblock


\bibitem[Matas et~al\mbox{.}(2018)]%
        {matas2018sim}
\bibfield{author}{\bibinfo{person}{Jan Matas}, \bibinfo{person}{Stephen James}, {and} \bibinfo{person}{Andrew~J Davison}.} \bibinfo{year}{2018}\natexlab{}.
\newblock \showarticletitle{Sim-to-real reinforcement learning for deformable object manipulation}. In \bibinfo{booktitle}{\emph{Conference on Robot Learning}}. PMLR, \bibinfo{pages}{734--743}.
\newblock


\bibitem[McGunnigle(2010)]%
        {mcgunnigle2010}
\bibfield{author}{\bibinfo{person}{G. McGunnigle}.} \bibinfo{year}{2010}\natexlab{}.
\newblock \showarticletitle{Detecting wet surfaces using near infrared lighting}.
\newblock \bibinfo{journal}{\emph{J. Opt. Soc. Am. A}} \bibinfo{volume}{27}, \bibinfo{number}{5} (\bibinfo{date}{May} \bibinfo{year}{2010}), \bibinfo{pages}{1137--1144}.
\newblock
\urldef\tempurl%
\url{https://doi.org/10.1364/JOSAA.27.001137}
\showDOI{\tempurl}


\bibitem[Pegram et~al\mbox{.}(2007)]%
        {pegram2007clinical}
\bibfield{author}{\bibinfo{person}{Anne Pegram}, \bibinfo{person}{Jacqueline Bloomfield}, {and} \bibinfo{person}{Anne Jones}.} \bibinfo{year}{2007}\natexlab{}.
\newblock \showarticletitle{Clinical skills: bed bathing and personal hygiene needs of patients}.
\newblock \bibinfo{journal}{\emph{British journal of nursing}} \bibinfo{volume}{16}, \bibinfo{number}{6} (\bibinfo{year}{2007}), \bibinfo{pages}{356--358}.
\newblock


\bibitem[Pratt and Williamson(1995)]%
        {pratt1995series}
\bibfield{author}{\bibinfo{person}{Gill~A Pratt} {and} \bibinfo{person}{Matthew~M Williamson}.} \bibinfo{year}{1995}\natexlab{}.
\newblock \showarticletitle{Series elastic actuators}. In \bibinfo{booktitle}{\emph{Proceedings 1995 IEEE/RSJ International Conference on Intelligent Robots and Systems. Human Robot Interaction and Cooperative Robots}}, Vol.~\bibinfo{volume}{1}. IEEE, \bibinfo{pages}{399--406}.
\newblock


\bibitem[Protsiv et~al\mbox{.}(2020)]%
        {10.7554/eLife.49555}
\bibfield{author}{\bibinfo{person}{Myroslava Protsiv}, \bibinfo{person}{Catherine Ley}, \bibinfo{person}{Joanna Lankester}, \bibinfo{person}{Trevor Hastie}, {and} \bibinfo{person}{Julie Parsonnet}.} \bibinfo{year}{2020}\natexlab{}.
\newblock \showarticletitle{Decreasing human body temperature in the United States since the Industrial Revolution}.
\newblock \bibinfo{journal}{\emph{eLife}}  \bibinfo{volume}{9} (\bibinfo{date}{jan} \bibinfo{year}{2020}), \bibinfo{pages}{e49555}.
\newblock
\showISSN{2050-084X}
\urldef\tempurl%
\url{https://doi.org/10.7554/eLife.49555}
\showDOI{\tempurl}


\bibitem[Sanchez et~al\mbox{.}(2018)]%
        {sanchez2018robotic}
\bibfield{author}{\bibinfo{person}{Jose Sanchez}, \bibinfo{person}{Juan-Antonio Corrales}, \bibinfo{person}{Belhassen-Chedli Bouzgarrou}, {and} \bibinfo{person}{Youcef Mezouar}.} \bibinfo{year}{2018}\natexlab{}.
\newblock \showarticletitle{Robotic manipulation and sensing of deformable objects in domestic and industrial applications: a survey}.
\newblock \bibinfo{journal}{\emph{The International Journal of Robotics Research}} \bibinfo{volume}{37}, \bibinfo{number}{7} (\bibinfo{year}{2018}), \bibinfo{pages}{688--716}.
\newblock


\bibitem[Saxen and Al-Hamadi(2014)]%
        {saxen2014color}
\bibfield{author}{\bibinfo{person}{Frerk Saxen} {and} \bibinfo{person}{Ayoub Al-Hamadi}.} \bibinfo{year}{2014}\natexlab{}.
\newblock \showarticletitle{Color-based skin segmentation: An evaluation of the state of the art}. In \bibinfo{booktitle}{\emph{2014 IEEE International Conference on Image Processing (ICIP)}}. IEEE, \bibinfo{pages}{4467--4471}.
\newblock


\bibitem[Sekachev et~al\mbox{.}(2020)]%
        {cvat}
\bibfield{author}{\bibinfo{person}{Boris Sekachev}, \bibinfo{person}{Nikita Manovich}, \bibinfo{person}{Maxim Zhiltsov}, \bibinfo{person}{Andrey Zhavoronkov}, \bibinfo{person}{Dmitry Kalinin}, \bibinfo{person}{Ben Hoff}, \bibinfo{person}{TOsmanov}, \bibinfo{person}{Dmitry Kruchinin}, \bibinfo{person}{Artyom Zankevich}, \bibinfo{person}{DmitriySidnev}, \bibinfo{person}{Maksim Markelov}, \bibinfo{person}{Johannes222}, \bibinfo{person}{Mathis Chenuet}, \bibinfo{person}{a andre}, \bibinfo{person}{telenachos}, \bibinfo{person}{Aleksandr Melnikov}, \bibinfo{person}{Jijoong Kim}, \bibinfo{person}{Liron Ilouz}, \bibinfo{person}{Nikita Glazov}, \bibinfo{person}{Priya4607}, \bibinfo{person}{Rush Tehrani}, \bibinfo{person}{Seungwon Jeong}, \bibinfo{person}{Vladimir Skubriev}, \bibinfo{person}{Sebastian Yonekura}, \bibinfo{person}{vugia truong}, \bibinfo{person}{zliang7}, \bibinfo{person}{lizhming}, {and} \bibinfo{person}{Tritin Truong}.} \bibinfo{year}{2020}\natexlab{}.
\newblock \bibinfo{booktitle}{\emph{opencv/cvat: v1.1.0}}.
\newblock
\urldef\tempurl%
\url{https://doi.org/10.5281/zenodo.4009388}
\showDOI{\tempurl}


\bibitem[Shetty and Ang(1996)]%
        {shetty1996active}
\bibfield{author}{\bibinfo{person}{Bharath~Ram Shetty} {and} \bibinfo{person}{Marcelo~H Ang}.} \bibinfo{year}{1996}\natexlab{}.
\newblock \showarticletitle{Active compliance control of a PUMA 560 robot}. In \bibinfo{booktitle}{\emph{Proceedings of IEEE International Conference on Robotics and Automation}}, Vol.~\bibinfo{volume}{4}. IEEE, \bibinfo{pages}{3720--3725}.
\newblock


\bibitem[Shimano et~al\mbox{.}(2017)]%
        {Shimano2017}
\bibfield{author}{\bibinfo{person}{Mihoko Shimano}, \bibinfo{person}{Hiroki Okawa}, \bibinfo{person}{Yuta Asano}, \bibinfo{person}{Ryoma Bise}, \bibinfo{person}{Ko Nishino}, {and} \bibinfo{person}{Imari Sato}.} \bibinfo{year}{2017}\natexlab{}.
\newblock \showarticletitle{Wetness and Color from a Single Multispectral Image}. In \bibinfo{booktitle}{\emph{2017 IEEE Conference on Computer Vision and Pattern Recognition (CVPR)}}. \bibinfo{pages}{321--329}.
\newblock
\urldef\tempurl%
\url{https://doi.org/10.1109/CVPR.2017.42}
\showDOI{\tempurl}


\bibitem[Siciliano and Villani(1999)]%
        {siciliano1999robot}
\bibfield{author}{\bibinfo{person}{Bruno Siciliano} {and} \bibinfo{person}{Luigi Villani}.} \bibinfo{year}{1999}\natexlab{}.
\newblock \bibinfo{booktitle}{\emph{Robot force control}}.
\newblock \bibinfo{publisher}{Springer Science \& Business Media}.
\newblock


\bibitem[Sorrentino and Remmert(2016)]%
        {sorrentino2016mosby}
\bibfield{author}{\bibinfo{person}{Sheila~A Sorrentino} {and} \bibinfo{person}{Leighann Remmert}.} \bibinfo{year}{2016}\natexlab{}.
\newblock \bibinfo{booktitle}{\emph{Mosby's Textbook for Nursing Assistants-E-Book}}.
\newblock \bibinfo{publisher}{Elsevier Health Sciences}.
\newblock


\bibitem[Stokes et~al\mbox{.}(2014)]%
        {doi:10.1089/soro.2013.0002}
\bibfield{author}{\bibinfo{person}{Adam~A. Stokes}, \bibinfo{person}{Robert~F. Shepherd}, \bibinfo{person}{Stephen~A. Morin}, \bibinfo{person}{Filip Ilievski}, {and} \bibinfo{person}{George~M. Whitesides}.} \bibinfo{year}{2014}\natexlab{}.
\newblock \showarticletitle{A Hybrid Combining Hard and Soft Robots}.
\newblock \bibinfo{journal}{\emph{Soft Robotics}} \bibinfo{volume}{1}, \bibinfo{number}{1} (\bibinfo{year}{2014}), \bibinfo{pages}{70--74}.
\newblock
\urldef\tempurl%
\url{https://doi.org/10.1089/soro.2013.0002}
\showDOI{\tempurl}
\showeprint{https://doi.org/10.1089/soro.2013.0002}


\bibitem[Sun et~al\mbox{.}(2019)]%
        {sun2019}
\bibfield{author}{\bibinfo{person}{Yuxiang Sun}, \bibinfo{person}{Weixun Zuo}, {and} \bibinfo{person}{Ming Liu}.} \bibinfo{year}{2019}\natexlab{}.
\newblock \showarticletitle{RTFNet: RGB-Thermal Fusion Network for Semantic Segmentation of Urban Scenes}.
\newblock \bibinfo{journal}{\emph{IEEE Robotics and Automation Letters}} \bibinfo{volume}{4}, \bibinfo{number}{3} (\bibinfo{year}{2019}), \bibinfo{pages}{2576--2583}.
\newblock
\urldef\tempurl%
\url{https://doi.org/10.1109/LRA.2019.2904733}
\showDOI{\tempurl}


\bibitem[Vasic and Billard(2013)]%
        {6630576}
\bibfield{author}{\bibinfo{person}{Milos Vasic} {and} \bibinfo{person}{Aude Billard}.} \bibinfo{year}{2013}\natexlab{}.
\newblock \showarticletitle{Safety issues in human-robot interactions}. In \bibinfo{booktitle}{\emph{2013 IEEE International Conference on Robotics and Automation}}. \bibinfo{pages}{197--204}.
\newblock
\urldef\tempurl%
\url{https://doi.org/10.1109/ICRA.2013.6630576}
\showDOI{\tempurl}


\bibitem[Villani et~al\mbox{.}(2013)]%
        {villani2013null}
\bibfield{author}{\bibinfo{person}{Luigi Villani}, \bibinfo{person}{Hamid Sadeghian}, {and} \bibinfo{person}{Bruno Siciliano}.} \bibinfo{year}{2013}\natexlab{}.
\newblock \showarticletitle{Null-space impedance control for physical human-robot interaction}. In \bibinfo{booktitle}{\emph{Romansy 19--Robot Design, Dynamics and Control: Proceedings of the 19th CISM-Iftomm Symposium}}. Springer, \bibinfo{pages}{193--200}.
\newblock


\bibitem[Ward et~al\mbox{.}(2017)]%
        {Ward2017}
\bibfield{author}{\bibinfo{person}{WH Ward}, \bibinfo{person}{F Lambreton}, {and} \bibinfo{person}{N~et~al. Goel}.} \bibinfo{year}{2017}\natexlab{}.
\newblock \bibinfo{booktitle}{\emph{Clinical Presentation and Staging of Melanoma. In: Ward WH, Farma JM, editors. Cutaneous Melanoma: Etiology and Therapy}}.
\newblock \bibinfo{publisher}{Brisbane (AU): Codon Publications}.
\newblock
\urldef\tempurl%
\url{https://www.ncbi.nlm.nih.gov/books/NBK481857/table/chapter6.t1/ doi: 10.15586/codon.cutaneousmelanoma.2017.ch6}
\showURL{%
\tempurl}


\bibitem[Xie et~al\mbox{.}(2021)]%
        {xie2021segformer}
\bibfield{author}{\bibinfo{person}{Enze Xie}, \bibinfo{person}{Wenhai Wang}, \bibinfo{person}{Zhiding Yu}, \bibinfo{person}{Anima Anandkumar}, \bibinfo{person}{Jose~M. Alvarez}, {and} \bibinfo{person}{Ping Luo}.} \bibinfo{year}{2021}\natexlab{}.
\newblock \bibinfo{title}{SegFormer: Simple and Efficient Design for Semantic Segmentation with Transformers}.
\newblock
\newblock
\showeprint[arxiv]{2105.15203}~[cs.CV]


\bibitem[Yukawa et~al\mbox{.}(2010)]%
        {Yukawa2010}
\bibfield{author}{\bibinfo{person}{Toshihiro Yukawa}, \bibinfo{person}{Nobuhiro Nakata}, \bibinfo{person}{Goro Obinata}, {and} \bibinfo{person}{Taira Makino}.} \bibinfo{year}{2010}\natexlab{}.
\newblock \showarticletitle{Assistance system for bedridden patients to reduce the burden of nursing care (first report — Development of a multifunctional electric wheelchair, portable bath, lift, and mobile robot with portable toilet)}. In \bibinfo{booktitle}{\emph{2010 IEEE/SICE International Symposium on System Integration}}. \bibinfo{pages}{132--139}.
\newblock
\urldef\tempurl%
\url{https://doi.org/10.1109/SII.2010.5708314}
\showDOI{\tempurl}


\bibitem[Zhang et~al\mbox{.}(2023)]%
        {zhang2023cmx}
\bibfield{author}{\bibinfo{person}{Jiaming Zhang}, \bibinfo{person}{Huayao Liu}, \bibinfo{person}{Kailun Yang}, \bibinfo{person}{Xinxin Hu}, \bibinfo{person}{Ruiping Liu}, {and} \bibinfo{person}{Rainer Stiefelhagen}.} \bibinfo{year}{2023}\natexlab{}.
\newblock \bibinfo{title}{CMX: Cross-Modal Fusion for RGB-X Semantic Segmentation with Transformers}.
\newblock
\newblock
\showeprint[arxiv]{2203.04838}~[cs.CV]


\bibitem[Zhou et~al\mbox{.}(2022)]%
        {Zhou2022}
\bibfield{author}{\bibinfo{person}{Heng Zhou}, \bibinfo{person}{Chunna Tian}, \bibinfo{person}{Zhenxi Zhang}, \bibinfo{person}{Qizheng Huo}, \bibinfo{person}{Yongqiang Xie}, {and} \bibinfo{person}{Zhongbo Li}.} \bibinfo{year}{2022}\natexlab{}.
\newblock \showarticletitle{Multispectral Fusion Transformer Network for RGB-Thermal Urban Scene Semantic Segmentation}.
\newblock \bibinfo{journal}{\emph{IEEE Geoscience and Remote Sensing Letters}}  \bibinfo{volume}{19} (\bibinfo{year}{2022}), \bibinfo{pages}{1--5}.
\newblock
\urldef\tempurl%
\url{https://doi.org/10.1109/LGRS.2022.3179721}
\showDOI{\tempurl}


\bibitem[Zhu et~al\mbox{.}(2022)]%
        {zhu2022challenges}
\bibfield{author}{\bibinfo{person}{Jihong Zhu}, \bibinfo{person}{Andrea Cherubini}, \bibinfo{person}{Claire Dune}, \bibinfo{person}{David Navarro-Alarcon}, \bibinfo{person}{Farshid Alambeigi}, \bibinfo{person}{Dmitry Berenson}, \bibinfo{person}{Fanny Ficuciello}, \bibinfo{person}{Kensuke Harada}, \bibinfo{person}{Jens Kober}, \bibinfo{person}{Xiang Li}, {et~al\mbox{.}}} \bibinfo{year}{2022}\natexlab{}.
\newblock \showarticletitle{Challenges and outlook in robotic manipulation of deformable objects}.
\newblock \bibinfo{journal}{\emph{IEEE Robotics \& Automation Magazine}} \bibinfo{volume}{29}, \bibinfo{number}{3} (\bibinfo{year}{2022}), \bibinfo{pages}{67--77}.
\newblock


\bibitem[Zlatintsi et~al\mbox{.}(2020)]%
        {zlatintsi2020support}
\bibfield{author}{\bibinfo{person}{Athanasia Zlatintsi}, \bibinfo{person}{AC Dometios}, \bibinfo{person}{Nikolaos Kardaris}, \bibinfo{person}{Isidoros Rodomagoulakis}, \bibinfo{person}{Petros Koutras}, \bibinfo{person}{X Papageorgiou}, \bibinfo{person}{Petros Maragos}, \bibinfo{person}{Costas~S Tzafestas}, \bibinfo{person}{Panagiotis Vartholomeos}, \bibinfo{person}{Klaus Hauer}, {et~al\mbox{.}}} \bibinfo{year}{2020}\natexlab{}.
\newblock \showarticletitle{I-Support: A robotic platform of an assistive bathing robot for the elderly population}.
\newblock \bibinfo{journal}{\emph{Robotics and Autonomous Systems}}  \bibinfo{volume}{126} (\bibinfo{year}{2020}), \bibinfo{pages}{103451}.
\newblock


\bibitem[Zlatintsi et~al\mbox{.}(2017)]%
        {zlatintsi2017}
\bibfield{author}{\bibinfo{person}{Athanasia Zlatintsi}, \bibinfo{person}{Isidoros Rodomagoulakis}, \bibinfo{person}{Petros Koutras}, \bibinfo{person}{A.~C. Dometios}, \bibinfo{person}{Vassilis Pitsikalis}, \bibinfo{person}{Costas~S. Tzafestas}, {and} \bibinfo{person}{Petros Maragos}.} \bibinfo{year}{2017}\natexlab{}.
\newblock \showarticletitle{Multimodal Signal Processing and Learning Aspects of Human-Robot Interaction for an Assistive Bathing Robot}.
\newblock \bibinfo{journal}{\emph{CoRR}}  \bibinfo{volume}{abs/1711.01775} (\bibinfo{year}{2017}).
\newblock
\showeprint[arXiv]{1711.01775}
\urldef\tempurl%
\url{http://arxiv.org/abs/1711.01775}
\showURL{%
\tempurl}


\end{thebibliography}
